\documentclass[conference]{IEEEtran}

\usepackage{times}
\usepackage{amsmath,amssymb,amsthm,xpatch}
\usepackage{xcolor}
\let\labelindent\relax
\usepackage{enumitem}
\usepackage{multicol}
\usepackage{graphicx,subcaption}
\usepackage{hyperref}
\hypersetup{
    colorlinks=true,
	citecolor=red!80,
    linkcolor=blue!80,
    filecolor=magenta,      
    urlcolor=cyan,
}

\usepackage[sort&compress,numbers]{natbib}
\usepackage{bm}

\usepackage{tikz} 
\usepackage{tkz-graph}
\usepackage{tkz-berge}
	\usetikzlibrary{backgrounds,fit,shapes,snakes,arrows,shapes.geometric,positioning}
	\usetikzlibrary{intersections,patterns,shapes.misc}
	\usetikzlibrary{decorations.pathmorphing}
	\tikzstyle{block} = [rectangle, rounded corners, minimum width=3cm, minimum height=1cm,text centered, draw=black, fill=red!30]
	\tikzstyle{new} = [rectangle, rounded corners, minimum width=1cm, minimum
	height=1cm,text centered, draw=black, fill=blue!10!white, dashed]
	\tikzstyle{arrow} = [thick,->,>=stealth]
	\usetikzlibrary{calc, quotes}
\usepackage{pgfplots}
\usepackage{xcolor}

\definecolor{kkGreen}{RGB}{201,232,206}
\definecolor{kkRed}{RGB}{255,196,215}
\definecolor{kkBlue}{RGB}{214,226,255}

\tikzset{hfit/.style={rounded rectangle, inner xsep=0pt, fill=#1!10},
           vfit/.style={rounded corners, fill=#1!30}}
\usepgfplotslibrary{fillbetween}

\usepackage{algorithmicx}
\usepackage{algorithm}
\usepackage[]{algpseudocode}
\algnewcommand\algorithmicinput{\textbf{Input:}}
\algnewcommand\INPUT{\item[\algorithmicinput]}
\algnewcommand{\LeftComment}[1]{\Statex #1}

\usepackage{fontawesome}
\usepackage{float}

\usepackage{mathtools}
\usepackage{icomma}
\usepackage{stmaryrd}

\usepackage{eucal}

\newcommand{\OPT}{\mathrm{OPT}}
\newcommand{\UPT}{\mathrm{UPT}}

\newcommand{\VV}{\mathcal{V}}

\newcommand{\EE}{{\mathcal{E}}}

\newcommand{\zero}{\mathbf{0}}

\newcommand{\HH}{\mathbb{I}}

\newcommand{\zz}{\boldsymbol{\ell}}

\newcommand{\ppp}{\boldsymbol\pi}
\newcommand{\AAA}{\mathbf{A}}

\newcommand{\Acal}{\mathcal{A}}

\newcommand{\Scal}{\mathcal{S}}

\newcommand{\Bcal}{\mathcal{B}}

\newcommand{\Wcal}{\mathcal{W}}
\newcommand{\Ccal}{\mathcal{C}}
\newcommand{\Fcal}{\mathcal{F}}

\newcommand{\JJ}{\mathbf{{J}}}

\DeclareMathAlphabet\mathbfcal{OMS}{cmsy}{b}{n}

\newcommand{\Rset}{\mathbb{R}}

\DeclareMathOperator*{\argmin}{arg\,min}

\DeclareMathOperator*{\argmax}{arg\,max}

\newcommand{\Ecal}{\mathcal{E}}
\newcommand{\Vcal}{\mathcal{V}}
\newcommand{\Lcal}{\mathcal{E}}

\let\oldsim\sim
\renewcommand{\sim}{\overset{\Lcal}{\oldsim}}

\newcommand{\ptext}[1]{\text{\footnotesize\color{black!70!white}#1}}
\newcommand{\edg}{\mathsf{edges}}
\newcommand{\OPTv}{\mathrm{OPT}_\text{v}}
\newcommand{\OPTe}{\mathrm{OPT}_\text{e}}
\newcommand{\fe}{f_\text{e}}
\newcommand{\fv}{f_\text{v}}
\newcommand{\vcover}{c_w}
\newcommand{\Vcover}{\mathcal{C}_w}

\newcommand{\Gcal}{\mathcal{G}_\mathsf{x}}

\newcommand{\Ucal}{\mathcal{U}}

\newcommand{\Qcal}{\mathcal{Q}}

\newcommand{\Vall}{\Vcal_\mathsf{x}}
\newcommand{\Eall}{\Lcal_\mathsf{x}}
\newcommand{\Vgrd}{\VV_\text{grd}}
\newcommand{\Egrd}{\Lcal_\text{grd}}
\newcommand{\cost}{\text{cost}}

\newcommand{\Vnew}{\VV_\text{new}}
\newcommand{\Vprev}{\VV_\text{prev}}

\newtheorem{lemma}{Lemma}
\newtheorem{theorem}{Theorem}

\theoremstyle{definition}
\newtheorem{definition}{Definition}

\newtheorem{remark}{Remark}

\xpatchcmd{\proof}{\noindent}{\noindent}{}{}

\usepackage{bbm}
\usepackage{mdframed}
\makeatletter
\def\NAT@def@citea{\def\@citea{\NAT@separator}}
\makeatother

\begin{document}

\title{\Large Near-Optimal Budgeted Data Exchange for Distributed Loop Closure Detection}

\author{\authorblockN{Yulun Tian\authorrefmark{1}, Kasra Khosoussi\authorrefmark{1}, Matthew Giamou\authorrefmark{2}, Jonathan P.~How\authorrefmark{1} and Jonathan Kelly\authorrefmark{2}}\vspace{0.15cm}
\authorblockA{\authorrefmark{1}Laboratory for Information and Decision Systems
(LIDS)\\
Massachusetts Institute of Technology
\\Email: \texttt{\{yulun,kasra,jhow\}@mit.edu}}
\authorblockA{\authorrefmark{2}Space \& Terrestrial Autonomous
Robotic Systems (STARS) Laboratory\\
University of Toronto Institute for Aerospace Studies (UTIAS)
\\Email: \texttt{\{matthew.giamou,jonathan.kelly\}@robotics.utias.utoronto.ca}}
}

\maketitle

\begin{abstract}
  Inter-robot loop closure detection is a core problem in collaborative SLAM
  (CSLAM). 
  Establishing inter-robot loop closures is a resource-demanding
  process,  
  during which robots must consume a substantial amount of
  mission-critical resources (e.g., battery and bandwidth)
  to exchange sensory
  data. However, 
  even with the most resource-efficient techniques,
  the resources available onboard may be insufficient for verifying
  every potential loop closure. This work addresses this critical challenge by proposing a 
  resource-adaptive framework for distributed loop closure detection.
  We seek to maximize task-oriented objectives subject to a budget constraint on total data transmission.
  This problem is in general NP-hard.
  We approach this problem from different perspectives and leverage existing results on
  monotone submodular maximization to provide efficient approximation algorithms
  with performance guarantees. 
  The proposed approach is extensively evaluated using the KITTI odometry benchmark
  dataset and synthetic Manhattan-like datasets. 
\end{abstract}

\IEEEpeerreviewmaketitle

\section{Introduction}
\label{sec: intro}

Multirobot systems can provide solutions to complex
large-scale tasks that are otherwise beyond the capability of a single robot.
Collaborative Simultaneous Localization and Mapping (CSLAM) is an indispensable
part of any multirobot system requiring robots to navigate in an unknown
\hbox{GPS-denied} environment; see \cite{saeedi2016multiple} for a recent
survey. {Inter-robot} loop closures lie at the heart of CSLAM: they tie
individual trajectories and maps together, and allow spatial information to flow
from one robot to the entire team.  As a result, the success of CSLAM---and,
consequently, that of the mission---is directly dependent upon 
detection of informative inter-robot loop closures.

Mobile robots are necessarily constrained by the limited resources
they can carry on-board (e.g., batteries, processing power, wireless radios). These
resources are especially scarce in the case of cost-effective multirobot
systems. Addressing these limitations is a
critical challenge in CSLAM. In particular, detecting inter-robot loop closures
is a resource-demanding task
\mbox{\cite{Giamou18_ICRA,CieslewskiS17,cieslewski2017efficient,CieslewskiChoudhary17,saeedi2016multiple}}
since robots must consume substantial amounts of energy and bandwidth to
exchange their observations.
This challenge has been recognized in numerous recent publications and has sparked a growing interest
in designing \emph{resource-efficient} 
frameworks that aim to minimize resource consumption due to data exchange in
CSLAM front-ends
\cite{choudhary2017,CieslewskiS17,cieslewski2017efficient,Giamou18_ICRA,leonardos2017distributed}.
However, 
even with the most \hbox{resource-efficient} data exchange policies \cite{CieslewskiS17,cieslewski2017efficient,CieslewskiChoudhary17,Giamou18_ICRA},
the available bandwidth and/or the allocated budget on energy
consumption may still be insufficient for verifying all \emph{potential} inter-robot loop
closures.
It is thus crucial for robots to be able to seamlessly
\emph{adapt} to the allocated resource budgets 
by selecting a budget-feasible subset of potential loop closures.

We propose a \emph{resource-adaptive} framework that aims to maximize
task-oriented monotone
submodular objectives subject to a
budget constraint on data transmission in CSLAM front-ends.
\citet{Giamou18_ICRA} have recently shown that the minimum amount of data transmission required for
verifying a given set of potential inter-robot loop closures is determined by
the size of the minimum 
vertex cover in the corresponding \emph{exchange graph}.
This immediately implies that verifying the budget-feasibility of a proposed solution to our problem
is basically equivalent to solving the vertex cover problem and thus is NP-complete in general. 
Furthermore, in this setting, the greedy measurement selection algorithms (i.e.,
selecting edges greedily)
designed for
standard cardinality constraints on the set of selected measurements
\cite{kasra18ijrr,carlone2017attention,shamaiah2010greedy} 
lack the necessary foresight and thus may exhaust their data transmission budgets
rapidly, leading to arbitrarily poor
performance. 

Despite these challenges, after a change of variables and by
establishing an approximation factor preserving reduction, we show that the natural greedy algorithm 
can still provide constant-factor approximation guarantees. 
The performance of the proposed approach is extensively validated using both real-world and simulated datasets. 
In particular, empirical near-optimality is demonstrated by leveraging natural
convex relaxations.
Ultimately, our work enables
rendezvousing robots to collaboratively search for inter-robot loop
closures by exchanging a near-optimal subset of their
observations without exceeding an allocated data transmission budget.

\vspace{0.1cm}
\noindent\emph{Notation.} 
Bold lower-case and upper-case letters are reserved for vectors and matrices, respectively.
$\AAA \succeq \zero$ (resp., $\AAA \succ \zero$) means that $\AAA$ is
positive semidefinite (resp., positive definite).
$\Vcover(\Lcal)$ denotes the minimum weighted vertex cover of the graph induced by 
the edge set $\Lcal$ where vertices are weighted by a positive weight function $w$.
Finally, the value of $\Vcover(\Lcal)$ is denoted by $\vcover(\Lcal) \triangleq \sum_{v
\in \Vcover(\Lcal)} w(v)$.

\section{Related Work}
\label{sec:related}

This paper focuses on CSLAM front-ends; see, e.g.,
\cite{saeedi2016multiple,choudhary2017,nerurkar2009distributed} and references therein for
state-of-the-art resource-efficient CSLAM back-ends. We limit our discussion 
to prior works that either propose resource-aware CSLAM front-ends or use methods similar to 
ours on related problems.

\citet{choudhary2017} circumvent the resource-intensive nature of data exchange
in CSLAM front-ends by utilizing object models to reduce communication between robots, effectively 
compressing sensory observations with high-level semantic labels. Needless to
say, their approach is limited to environments filled with known objects.

\citet{CieslewskiS17,cieslewski2017efficient} 
focus on designing data-efficient and scalable schemes for the ``online query''
phase of distributed place recognition.
During this phase,
robots exchange compact representations of their observations 
in order to decide to whom to send a full place query. 
The proposed methods reduce the amount of data transmission in this phase
from $O(n^2)$ to
$O(n)$ where $n$ is the number of robots.
It is empirically shown that the applied heuristics only incur a 
small loss in place recognition recall.
We, however, focus on resource consumption due to 
exchanging \emph{full} observations that takes place \emph{after} this initial phase. 
As a result,
our resource-adaptive framework can be used alongside
\cite{cieslewski2017efficient,CieslewskiS17,CieslewskiChoudhary17}
to improve the overall resource-efficiency of 
the system.

Once the query phase is complete, robots need to exchange full image
keypoints for geometric verification and relative pose estimation.
\citet{Giamou18_ICRA} focus on resource-optimal exchange policies for this stage. The proposed problem formulation takes into account both the
total amount of exchanged data and the induced division of labor among robots.
The authors show that in the pairwise case, this problem is essentially
equivalent to the 
minimum bipartite vertex cover problem and hence can be solved
efficiently. Our paper is similar to \cite{Giamou18_ICRA} in that we also consider
the more communication-intensive phase of full data exchange. However, our approach also
explicitly considers the ``value'' of (e.g., information gained by) sharing each observation, and is able
to deal with hard communication budgets.

Belief pruning and graph sparsification constitute another line of research in
resource-efficient SLAM \cite{kretzschmar2012, paull2015communication,
Vial2011, Huang2013}. 
In particular, \citet{paull2015communication} consider CSLAM in 
underwater environments where acoustic communication is high-latency,
low-bandwidth, and unreliable. A consistent sparsification scheme is proposed in
\cite{paull2015communication} to reduce the amount of
transmitted data after marginalizing out variables. In \cite{paull2016unified}, the authors further look at the
combined problem of variable selection and graph sparsification, and propose a
unified framework that takes into account computation, communication, and memory
budgets. 
In this paper, we also cast our
problem in a similar resource-adaptive setting. Unlike
\cite{paull2016unified}, however, we focus on the complementary problem of
inter-robot loop closure detection in CSLAM front-ends.

\citet{kasra16wafr,kasra18ijrr} and \citet{carlone2017attention} have proposed
measurement selection schemes for SLAM and 
visual-inertial navigation (VIN).
In both applications,
a cardinality constraint is imposed 
on the selected measurements
to control the computational cost of solving the underlying 
estimation problem.
In \cite{kasra18ijrr,kasra16wafr}, 
the authors use weighted tree-connectivity \cite{kasra16icra} as a
graphical surrogate for the \mbox{D-optimality} criterion 
to select high quality measurements. 
Similarly, 
in the
context of VIN, 
\citet{carlone2017attention} 
use the D-optimality and E-optimality criteria to select a ``valuable'' subset of visual features for localization.
Both of these approaches
\cite{kasra16wafr,kasra18ijrr,carlone2017attention} leverage the monotone
submodular property of the D-criterion to provide performance guarantees. In
this paper, we also use the objective functions introduced in
\cite{kasra18ijrr,kasra16wafr,carlone2017attention}, albeit for
\emph{communication-constrained} scenarios.

\section{Problem Formulation:\\Opt. Resource-Adaptive Loop Closure Detection}
\label{sec:problem_formulation}

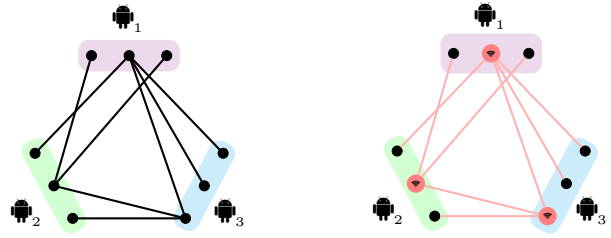
\begin{figure}[t]
  \centering
  \hspace{-0.04\textwidth}
  \begin{subfigure}[t]{0.2\textwidth}
	\centering
	\begin{tikzpicture}[scale=1]
	\tikzstyle{vertex}=[circle,fill,scale=0.4,draw]
	\tikzstyle{special vertex}=[circle,fill=red,scale=0.4,draw]
	\tikzstyle{square vertex}=[rectangle,fill,scale=0.5,draw]
	\tikzstyle{diamond vertex}=[regular polygon,regular polygon
	sides=3,rotate=45,fill,scale=0.3,draw]
	\node[vertex] at (1,2.598076) (a1) {};
	\node[vertex] at (1.5,2.598076) (a2) {};
	\node[vertex] at (2,2.598076) (a3) {};
	\node[vertex] at (0.25,1.299037) (b1) {};
	\node[vertex] at (0.5,0.866025) (b2) {};
	\node[vertex] at (0.75, 0.433013) (b3) {};
	\node[vertex] at (2.25,0.433013) (c1) {};
	\node[vertex] at (2.5,0.866025) (c2) {};
	\node[vertex] at (2.75,1.299037) (c3) {};
	%%%%%%%
	\node (r1) at (1.5,3.098076) [] {\faAndroid$_{_1}$};
	\node (r2) at (0.1464466,0.51247) [] {\faAndroid$_{_2}$};
	\node (r3) at (2.85355, 0.51247) [] {\faAndroid$_{_3}$};
	%%%%%%%
	\draw[thick](a1) -- (b2);
	\draw[thick](a2) -- (b1);
	\draw[thick](a2) -- (c2);
	\draw[thick](a2) -- (c3);
	\draw[thick](b2) -- (c1);
	\draw[thick](c1) -- (a2);
	\draw[thick](b2) -- (a3);	  
	\draw[thick](b3) -- (c1);
	\begin{pgfonlayer}{background}
		\node[fit=(a1)(a2)(a3),rounded corners,fill=violet!15,inner xsep=3pt,
inner ysep=4pt] {};
		\node[fit=(b1)(b2)(b3),rounded corners,fill=green!18,inner xsep=-4pt,
inner ysep=5pt,
		rotate=28] {};
		\node[fit=(c1)(c2)(c3),rounded corners,fill=cyan!18,inner xsep=-4pt,
inner ysep=4pt,
		rotate=152] {};
	\end{pgfonlayer}
	\end{tikzpicture}
	\caption{Example exchange graph}
	\label{fig:gex}
  \end{subfigure}\hspace{0.03\textwidth}
  ~
  \begin{subfigure}[t]{0.23\textwidth}
	\centering
	\begin{tikzpicture}[scale=1]
	\tikzstyle{vertex}=[circle,fill,scale=0.4,draw]
	\tikzstyle{special vertex}=[circle,fill=red!50,scale=0.3]
	\tikzstyle{square vertex}=[rectangle,fill,scale=0.5,draw]
	\tikzstyle{diamond vertex}=[regular polygon,regular polygon
	sides=3,rotate=45,fill,scale=0.3,draw]
	\node[vertex] at (1,2.598076) (a1) {};
	\node[special vertex] at (1.5,2.598076) (a2) {\faWifi};
	\node[vertex] at (2,2.598076) (a3) {};
	\node[vertex] at (0.25,1.299037) (b1) {};
	\node[special vertex] at (0.5,0.866025) (b2) {\faWifi};
	\node[vertex] at (0.75, 0.433013) (b3) {};
	\node[special vertex] at (2.25,0.433013) (c1) {\faWifi};
	\node[vertex] at (2.5,0.866025) (c2) {};
	\node[vertex] at (2.75,1.299037) (c3) {};
	%%%%%%%
	\node (r1) at (1.5,3.098076) [] {\faAndroid$_{_1}$};
	\node (r2) at (0.1464466,0.51247) [] {\faAndroid$_{_2}$};
	\node (r3) at (2.85355, 0.51247) [] {\faAndroid$_{_3}$};
	%%%%%%%
	\draw[thick,red!30](a1) -- (b2);
	\draw[thick,red!30](a2) -- (b1);
	\draw[thick,red!30](a2) -- (c2);
	\draw[thick,red!30](a2) -- (c3);
	\draw[thick,red!30](b2) -- (c1);
	\draw[thick,red!30](c1) -- (a2);
	\draw[thick,red!30](b2) -- (a3);	  
	\draw[thick,red!30](b3) -- (c1);
	\begin{pgfonlayer}{background}
		\node[fit=(a1)(a2)(a3),rounded corners,fill=violet!15,inner xsep=3pt,
inner ysep=4pt] {};
		\node[fit=(b1)(b2)(b3),rounded corners,fill=green!18,inner xsep=-4pt,
inner ysep=5pt,
		rotate=28] {};
		\node[fit=(c1)(c2)(c3),rounded corners,fill=cyan!18,inner xsep=-4pt,
inner ysep=4pt,
		rotate=152] {};
	\end{pgfonlayer}
	\end{tikzpicture}
	\caption{A lossless exchange policy}
	\label{fig:policy}
  \end{subfigure}
  \caption{\small (a) An example exchange graph $\Gcal$ in a $3$-rendezvous where each
	robot owns three vertices (observations). To verify a 
  potential inter-robot loop 
  closure between two connected vertices, at least one robot needs to share 
  its observation with the other robot. (b) A 
  lossless exchange policy in which the observations associated to the vertices
  marked in red are 
  transmitted. In the optimal exchange policy, robots must exchange $3$
  observations to cover all potential loop closures. Now if robots are only
  permitted to exchange at most $b = 2$
observations, they must decide which budget-feasible subset of potential loop closures is most
valuable (in expectation) based a task-oriented objective.}
  \label{fig:diagram}
\end{figure}

We begin by reviewing the general structure of decentralized inter-robot loop
closure detection schemes \cite{Giamou18_ICRA,cieslewski2017efficient,CieslewskiS17}.
CSLAM, by definition, relies on communication. Although maintaining a fully
connected communication
network may not be feasible at all times, robots must be able to
at least occasionally communicate with some of their peers during
close encounters---preplanned or otherwise. 
The inter-robot loop
closure detection process occurs during a \emph{rendezvous}. 
\begin{definition}[Rendezvous]
  \label{def:rendezvous}
  An $n$-\emph{rendezvous} ($n \geq 2$) refers to the situation where 
  $n$ robots are positioned such that each of them will receive the data broadcasted by
  any other robot in that group owing to the broadcast nature of wireless medium.
\end{definition}

\subsection{Metadata Exchange}
Each robot arrives at a rendezvous with its own unique collection of observations
(e.g., images or laser scans) acquired throughout its mission.
Robots then exchange a \emph{compact representation} of their observations
(``metadata'' according to \cite{Giamou18_ICRA}) in order
to identify \emph{potential} loop closures. Several choices are 
bag-of-words vectors \cite{cieslewski2017efficient,Giamou18_ICRA},
low-dimensional feature vectors computed by a deep neural network
\cite{CieslewskiS17}, and/or spatial clues (i.e., estimated location with
uncertainty) if robots have already established a common reference frame
\cite{Giamou18_ICRA}. 
Comparing metadata reveals a number of potential matches. Each
potential match comes with a similarity score \cite{GalvezTRO12} or an
\emph{occurrence probability} (estimated directly or obtained by normalizing the
similarity scores) that estimates the
likelihood 
of that potential match corresponding to a true loop closure.
The result can be naturally represented as an \emph{exchange graph}
\cite{Giamou18_ICRA}.

\begin{definition}[Exchange Graph]
	\label{def:exchange_graph}
	An exchange graph \cite{Giamou18_ICRA} is a simple undirected
	graph $\Gcal = (\Vall,\Eall)$ where each vertex $v \in \Vall$ corresponds to
	an observation, and each edge $\{u,v\} \in \Eall$ corresponds to a potential
	inter-robot loop closure between the corresponding observations.
	Furthermore, $\Gcal$ is endowed with $w : \Vall \to
	\Rset_{>0}$ and $p : \Eall \to (0,1]$ that quantify the size of each
	observation,\footnote{For instance, this can be the number of keypoints in an
	  image.}
	and the occurrence probability of an edge, respectively.
	We make the simplifying assumption that edges occur independently.
\end{definition}
	Note that in an $n$-rendezvous, the exchange graph will be an $n$-partite
	graph. Figure \ref{fig:gex} illustrates a simple exchange graph with $n = 3$.
	For simplicity, we assume that a single robot 
	participating the rendezvous (known as the ``broker'' in \cite{Giamou18_ICRA})
	is responsible for
	forming the exchange graph and solving the subsequent optimization problem introduced in Section~\ref{subsec:two_perspectives}.
	A fully decentralized approach is left for future work.

\subsection{Optimal Lossless Data Exchange}
\label{sec:optLossless}
After identifying potential matches, robots must exchange full observations for geometric
verification and relative pose estimation \cite{Giamou18_ICRA}. 
This stage is substantially more communication-intensive than the metadata exchange phase as it involves the transmission of
actual observations (e.g., 
the complete set of keypoints in an image).
In
\cite{CieslewskiS17,cieslewski2017efficient}, each robot sends its full query
\emph{only} to the owner of the ``most promising'' potential match.
This heuristic obviously comes at the risk of losing some loop closures.

Our recent work \cite{Giamou18_ICRA} presents an efficient algorithm for finding the optimal
data exchange policy that minimizes data transmission among all \emph{lossless} (i.e., allows robots to verify
\emph{every} potential
match) exchange policies in $2$-rendezvous.
In \cite{Giamou18_ICRA} we demonstrate that by forming
the exchange graph and exploiting its unique structure one can come up with intelligent
resource-efficient data exchange policies which were not possible in
\cite{CieslewskiS17,cieslewski2017efficient}.
In particular, for $2$-rendezvous we prove that 
in the optimal exchange policy,
agents share the observations that correspond to the minimum weighted vertex
cover of the exchange graph, $\Vcover(\Eall) \subset \Vall$ \cite{Giamou18_ICRA}. 
Thus, the
minimum amount of data transmission required for verifying all potential matches is
given by $\vcover(\Eall) \triangleq \sum_{v \in \Vcover(\Eall)} w(v)$.

Here we note that this result can be trivially generalized to the case of
$n$-rendezvous.
Figure~\ref{fig:policy} illustrates an example.
The key difference between $2$-rendezvous and $n$-rendezvous is that finding
the optimal lossless exchange
policy (i.e., the minimum weighted vertex cover) is NP-hard for general
$n$-rendezvous (i.e., general $n$-partite graphs) when $n \geq 3$. Nevertheless,
by rounding the solution of linear programming (LP) relaxation we obtain a
lossless exchange policy with at most $2\, \vcover(\Eall)$ data
transmission \cite{vazirani2013approximation}.

\subsection{Data Exchange under Budgeted Communication}
\label{subsec:two_perspectives}
Intrinsic limitations of the communication channel (e.g., bandwidth in
underwater acoustic communication \cite{paull2015communication}), operational constraints, and/or resource scheduling policies (e.g., enforced to
regulate resource consumption) necessitate and lead to budgeted communication.
We are specifically interested in situations where the data transmission budget
$b$ is strictly less than $\vcover(\Eall)$. Based on our earlier remarks, one can
immediately conclude that in this regime, verifying all potential matches
without exceeding the budget $b$ becomes impossible. Therefore, resource-adaptation and
task-oriented prioritization are inevitable.
In what follows, we approach this 
problem from two perspectives. It will become clear shortly that these two
seemingly different viewpoints lead to two
problem statements that are intimately connected.

\vspace{0.2cm}

\subsubsection{Edge Selection} From the perspective of measurement selection \cite{kasra16wafr,kasra18ijrr,carlone2017attention}, 
we seek to select a subset of potential loop closures such that the minimum
amount of data transmission needed to verify them is at most $b$.
We need a task-oriented objective $\fe: 2^{\Eall} \to \Rset_{\geq 0}$ that quantifies the value
of all subsets of potential loop closures. Note that $\fe$ must take into account
the stochastic nature of potential matches, captured by the edge weights $p :
\Eall \to (0,1]$ in $\Gcal$. Later in this
section we discuss several suitable choices of $\fe$ for CSLAM and place
recognition. 
This perspective leads to the following problem statement:
\begin{equation}
\underset{\Lcal \subseteq \Eall}{\mathrm{maximize }}
\,\, \fe(\Lcal) \,\,\mathrm{s.t.}\,\,
\vcover(\Lcal) \leq b,
\label{eq:adapt}
\tag{$\mathtt{P}_\text{e}$}
\end{equation}
where $\vcover(\Lcal)$ gives the minimum amount of data transmission needed to
verify $\Lcal$ (see Section~\ref{sec:optLossless}) and $\vcover(\Lcal) \leq b$ imposes a
global communication budget on the rendezvousing robots.
\begin{remark}
  \label{rem:npcomplete}
  Merely deciding whether a given $\Lcal \subseteq \Eall$ is \ref{eq:adapt}-feasible,
  (i.e., there exists a vertex cover for $\Lcal$ with a size of at most
  $b$), is an instance of the weighted vertex cover problem
  which is NP-complete \cite{karp1972reducibility} in $n$-rendezvous with
  $n \geq 3$.
  Put differently, deciding whether a set of potential loop closures can be
  verified by exchanging at most $b$ units of data is NP-complete when $n \geq 3$.
\end{remark}

\subsubsection{Vertex Selection}
From the perspective of data exchange polices, one may naturally search for an
optimal subset of vertices (observations) that need to be broadcasted without
exceeding the budget $b$. Recall that once an observation is broadcasted, other
robots can collectively verify \emph{all} potential loop closures involving that observation
(i.e., incident to the selected vertex); see Figure~\ref{fig:diagram} and
\cite{Giamou18_ICRA}.
This motivates the following objective function,
\begin{align}
  \fv : 2^{\Vall} &\to \Rset_{\geq 0 }: \VV \mapsto
  \fe\big(\edg(\VV)\big).
  \label{eq:fvdef}
\end{align}
where $\edg : 2^{\Vall} \to 2^{\Eall}$ gives the set of all edges 
such that for each edge at least one end is in the given subset of vertices.
This intuitively means that the value of any subset of vertices
is equal to the $\fe$-value of all edges incident to those vertices. 
This leads to the following optimization problem:
\begin{equation}
	 \underset{\VV \subseteq \Vall}{\mathrm{maximize }}\,\,
	  \fv(\VV) \,\,	 \mathrm{\,\,s.t.\,\,} \sum_{v \in \VV} w(v) \leq b,
	  \tag{$\mathtt{P}_\text{v}$}
  \label{prob:pv}
\end{equation}
Note that the budget constraint in \ref{prob:pv} is simpler than what
we encountered in \ref{eq:adapt}. Needless to say, \ref{eq:adapt} and \ref{prob:pv} are closely related. 
The connection between these two problems is discussed and exploited in Section~\ref{sec:algorithm}.

\subsection{Objective Function}
\label{subsec:obj}
\begin{definition}
  \label{def:NMS}
  A set function $f: 2^{\Wcal} \to \Rset_{\geq 0}$ for a finite $\Wcal$ is normalized, monotone, and submodular (NMS) if it satisfies the following properties:
  \begin{itemize}
	\item[$\diamond$] Normalized: $f(\varnothing) = 0$.
  \item[$\diamond$] Monotone: for any $\Acal \subseteq \Bcal$, $f(\Acal) \leq f(\Bcal)$.
  \item[$\diamond$] Submodular: for any $\Acal \subseteq \Wcal$ and $\Bcal \subseteq \Wcal$,
  \\ $f(\Acal) + f(\Bcal) \geq 
  f(\Acal \cup \Bcal ) + f(\Acal \cap \Bcal)$. 
  \end{itemize}
\end{definition}

We focus only on cases where $\fe$ is NMS. 
In what follows, we briefly review two examples of such functions used recently for
measurement selection in the context of SLAM and VIN
\cite{kasra18ijrr,kasra16wafr,carlone2017attention}. In addition, we consider
also a third objective 
that is suitable
for general place recognition scenarios.
Note that our framework is compatible with \emph{any} NMS
objectives, and is not limited to the instances considered below.

\subsection*{1) D-optimality Criterion}
\label{subsec:FIM}
The D-optimality design criterion (D-criterion), defined as the log-determinant of
the Fisher information matrix (FIM), is one of the most popular design
criteria in the theory of optimal experimental design
with well-known geometrical and information-theoretic interpretations; see,
e.g., \cite{Joshi2009,Pukelsheim1993}.
D-criterion has been widely adopted in many
problems including sensor selection \cite{Joshi2009,shamaiah2010greedy} and
measurement selection in SLAM
\cite{carlone2017attention,kasra16wafr}. 
Let $\HH_\text{init} \succ \zero$ denote the information matrix of the joint CSLAM problem
before incorporating the potential loop closures. Moreover, let $\HH_e = \JJ_e^\top
\boldsymbol\Sigma_e^{-1} \JJ_e
\succeq \zero$ be
the information matrix associated to the candidate loop closure $e \in
\Eall$ in which $\JJ_e$ and $\boldsymbol\Sigma$ denote the 
measurement Jacobian matrix and the covariance of Gaussian noise, respectively. Following \cite{carlone2017attention}, one can approximate the
expected gain in the D-criterion as:
\begin{equation}
  f_{\text{FIM}}(\Lcal) \triangleq \log\det \Big(\HH_\text{init} + \sum_{e \in \Lcal}
  p(e) \cdot   \HH_e\Big) - \log\det \HH_\text{init}.
  \label{eq:FIM}
\end{equation}
$f_{\text{FIM}}$ is NMS \cite{shamaiah2010greedy,carlone2017attention}.

\subsection*{2) Tree-Connectivity}
\label{subsec:WST}
The D-criterion in 2D pose-graph SLAM can be closely approximated by the weighted number of
spanning trees (WST)---hereafter, tree-connectivity---in the graphical representation of SLAM
\cite{kasra16icra}. 
In \cite{kasra16wafr,kasra18ijrr} tree-connectivity is used as a graphical
surrogate for the D-criterion for (potential) loop closure selection.
Evaluating tree-connectivity is computationally cheaper than evaluating the
D-criterion and, furthermore, does not require any metric knowledge of robots' trajectories.
Let $t_{w_p}(\Ecal)$ and $t_{w_\theta}(\Ecal)$ denote the weighted number of
spanning trees in a pose-graph specified by the edge set
$\Ecal$ whose edges are weighted by the precision of the translational and rotational
measurements, respectively \cite{kasra18ijrr}. Furthermore, let
$\Ecal_\text{init}$ be the set of edges that exist in the CSLAM
pose-graph prior to the rendezvous. 
Define $\Phi(\Lcal) \triangleq
   2\cdot \log \mathbb{E}\big[t_{w_p}(\Ecal_\text{init} \cup \EE)\big] + \log
  \mathbb{E}\big[t_{w_\theta}(\Ecal_\text{init} \cup \EE)\big]$ where
  expectation is with respect to the anisotropic random graph model defined in
  Definition~\ref{def:exchange_graph}; see \cite{kasra18ijrr}.
\citet{kasra18ijrr} then seek to maximize the following objective:
\begin{align}
  {f}_{\text{WST}}(\Lcal) \triangleq
  \Phi(\Lcal) - \Phi(\varnothing).
  \label{eq:WST}
\end{align}
It is shown in \cite{kasra18ijrr,kasra16wafr} that $f_\text{WST}$
is NMS if the underlying pose-graph is connected prior to the rendezvous.
\subsection*{3) Expected Number of True Loop Closures}
\label{subsec:NLC} The previous two estimation-theoretic objective functions are well suited for
CSLAM. However, in the context of distributed place recognition, one may simply wish to
maximize the expected number of true loop closures (NLC) between the agents.
From Definition~\ref{def:exchange_graph} recall that the probability associated to a potential loop closure
$e \in \Eall $ is $p(e)$. Consequently, the expected
number of true loop closures in a subset of edges $\Lcal$ can be expressed as:
\begin{equation}
  f_{\text{NLC}}(\Lcal) \triangleq  \begin{cases}
\sum_{e \in
  \Lcal} p(e) & \Ecal \neq \varnothing,\\
  0 & \Ecal = \varnothing.
  \end{cases}.  \label{eq:NLC} 
\end{equation}
$f_\text{NLC}$ is clearly NMS.

\section{Algorithm and Theoretical Guarantees}
\label{sec:algorithm}

This section presents approximation algorithms for our two perspectives \ref{prob:pv} and
\ref{eq:adapt} with performance guarantees
that hold for any NMS $\fe$ and the corresponding $\fv$.

\begin{theorem}
  \normalfont
  For any NMS $\fe$, the corresponding $\fv$ is NMS.
  \label{thm:fe2fv}
\end{theorem}
This theorem implies that \ref{prob:pv} is an instance
of the classical problem of maximizing an NMS function subject to a knapsack
constraint. Although this class of problems generalizes the maximum coverage
problem \cite{hochbaum1996approximation} and thus are NP-hard in general, they enjoy a rich
body of results on constant-factor approximation algorithms
\cite{krauseSurvey}.\footnote{Recall that an $\alpha$-approximation algorithm for a maximization problem is an efficient
algorithm that produces solutions with a value of at least $\alpha\cdot\mathrm{OPT}$ for a constant $\alpha \in (0,1)$.}
We discuss these algorithms in more detail and show how they can be applied to
\ref{prob:pv} in Section~\ref{subsec:apxalg}.

Now, we show that by establishing a simple approximation factor
preserving reduction from \ref{eq:adapt} to \ref{prob:pv}, we can also
obtain constant-factor approximation schemes for \ref{eq:adapt}. 
Let $\OPTe$ and
$\OPTv$ be the optimal values of \ref{eq:adapt} and \ref{prob:pv},
respectively.
The following lemmas shed more light on the connection between our two
perspectives \ref{eq:adapt} and \ref{prob:pv}.

\begin{lemma}
  \normalfont
  \leavevmode\\[0.1cm]
  \vspace{-0.5cm}
  \begin{enumerate}
	\item For any \ref{prob:pv}-feasible ${\VV}$,
	  ${\edg(\VV)}$ is \ref{eq:adapt}-feasible.
	\item $\OPTe = \OPTv$. 
  \end{enumerate}
  \label{th:opts}
\end{lemma}
Using the above lemma, we establish the following approximation factor preserving reduction from 
\ref{eq:adapt} to \ref{prob:pv}.
It is worth mentioning this reduction belongs to a strong class known as 
\mbox{``S-reductions''} \cite{crescenzi1997short,
ausiello2012complexity}.

\begin{theorem}
  \normalfont
  Given an $\alpha$-approximation algorithm for
  \ref{prob:pv}, the following is
  an $\alpha$-approximation algorithm for \ref{eq:adapt}:
  \begin{enumerate}
	\item Run the $\alpha$-approximation algorithm on \ref{prob:pv} to produce
	  $\tilde{\VV}$.
	\item Return $\edg(\tilde{\VV})$.
  \end{enumerate}
  \label{th:alphaalg}
\end{theorem}
Theorem~\ref{th:alphaalg} illustrates how constant-factor approximation
algorithms for \ref{prob:pv} can be used as a proxy to obtain constant-factor approximation
algorithms for the dual perspective in \ref{eq:adapt}.
In Section~\ref{subsec:apxalg}, we show how the interplay between \ref{prob:pv}
and \ref{eq:adapt} can be exploited in certain situations to further improve the
approximation guarantees of our algorithms.

\subsection{Approximation Algorithms}
\label{subsec:apxalg}
\begin{table*}
  \begin{center}
	\caption{A summary of approximation algorithms for maximizing NMS functions
	  based on \cite{krauseSurvey}} \label{tab:results}
	\def\arraystretch{1.2}
	\begin{tabular}{ccc}
	  \hline
	  \textbf{\small Observation Size (Vertex Weight)}& \textbf{\footnotesize S-reduction to NMS Max. Under} & \textbf{\small Approximation Factor}\\
	  \hline
	  \hline
	  \small Non-uniform
	  & \small Knapsack constraint &
	  \small Greedy$\ast$ \cite{leskovec2007cost} $1/2 \cdot (1-1/e)$ \textbf{AND} $1-1/e$ \cite{sviridenko2004note} \\
	  \hline
	  \small Uniform &
	  \small Cardinality constraint & \small Standard Greedy $1-1/e$ \cite{nemhauser1978analysis}\\
	  \hline
	\end{tabular}
  \end{center}
\end{table*}
So far, we have shown that near-optimal solutions to
\ref{prob:pv} and---by virtue of Theorem~\ref{th:alphaalg}---\ref{eq:adapt}
can be obtained 
using constant-factor approximation algorithms for maximizing 
NMS functions under a knapsack constraint.
Some of these algorithms are listed
in Table~\ref{tab:results}; see \cite{krauseSurvey} for a comprehensive survey.
The approximation factors obtained by these algorithms hold for any NMS objectives. 
In the special case of maximizing the expected number of loop closures $f_\text{NLC}$ 
(Section~\ref{subsec:NLC})
under a cardinality constraint, 
\ref{prob:pv} reduces to the well-studied 
maximum coverage problem over a graph \cite{hochbaum1996approximation}. 
In this case, a simple procedure based on pipage rounding can improve the approximation factor to $3/4$ \cite{ageevGraphMaxCover}. Furthermore,
if the graph is bipartite, a specialized algorithm can improve the approximation factor to $8/9$ \cite{caskurluBipartiteMaxCover}.
Nonetheless, in this work we focus on greedy algorithms 
described in Table~\ref{tab:results}
due to their generality,
computational efficiency, and incremental nature (see Remark~\ref{rem:anycomm}).

In many real-world scenarios, the number of
primitives (e.g., keypoints in an image) is roughly the same
across all measurements. By ignoring insignificant 
variations in observations sizes, one can assume that
each vertex (i.e., observation) has unit weight or size. In this case, the knapsack constraint
in \ref{prob:pv} reduces to a cardinality constraint $|\VV|\leq b$.
We first discuss this case and then revisit the more general case of
knapsack constraints.\vspace{0.4cm}

\noindent1) \underline{\it Uniform Observation Size}:
The standard greedy algorithm gives the optimal approximation factor for
maximizing general NMS functions under a cardinality constraint (Table~\ref{tab:results}). This algorithm, when applied on \ref{prob:pv} under
a cardinality constraint, is as follows: for $b$ rounds, greedily pick (without
replacement) a vertex $v$
with the highest marginal gain $\fv(\VV \cup \{v\}) - \fv(\VV)$, where
$\VV$ denotes the current set of selected vertices. 
Refer to Algorithm~\ref{alg:vgreedy} in the appendix for the complete pseudocode.
The solution $\VV_\text{grd}$ produced by the greedy algorithm satisfies 
$\fv(\VV_\text{grd}) \geq (1-1/e)\cdot\OPTv$ \cite{nemhauser1978analysis}.
Furthermore, by our approximation-factor-preserving reduction (Theorem~\ref{th:alphaalg}), 
$\VV_\text{grd}$ can in 
turn be mapped to a \ref{eq:adapt}-feasible subset of edges $\Lcal_\text{grd} = \edg(\VV_\text{grd})$, for which we also have
$\fe(\Lcal_\text{grd}) \geq (1-1/e)\cdot\OPTe$.

Remarkably, in some cases the interplay between \ref{eq:adapt} and \ref{prob:pv} reveals a pathway to further improve the solution 
even after $b$ rounds of standard greedy selection. 
We already know that $\Lcal_\text{grd}$ can be covered by
a subset of vertices of size $b$, namely $\VV_\text{grd}$. However,
there may be an even ``cheaper'' (i.e., of size $b^\prime < b$) subset of vertices that covers the entire
$\Lcal_\text{grd}$. 
Such a subset can be found by computing the minimum vertex cover of the graph induced by
$\Lcal_\text{grd}$. 
This is in general NP-hard.
We can, however, find a vertex cover by rounding a solution of the LP relaxation of this problem; see \cite{vazirani2013approximation}. For
$2$-rendezvous (bipartite exchange graphs), this gives the minimum vertex cover
as noted in \cite{Giamou18_ICRA}. As mentioned earlier, for general
\mbox{$n$-rendezvous} ($n\geq 3$), size of the resulting vertex cover is guaranteed to be at most
twice the size of the minimum vertex cover; see,
e.g., \cite{vazirani2013approximation}.
If such a subset can be found, we can continue running the greedy algorithm for
$b-b^\prime$ additional rounds while still ensuring that the final solution is
budget-feasible.  Suppose repeating the process of recomputing the
(approximate) minimum vertex cover leads to $b+k$ rounds of greedy
decisions in total and produces $\VV_\text{grd}$. 
Using \cite[Theorem 1.5]{krauseSurvey}, it can be shown that 
$\fv(\VV_\text{grd}) \geq (1-1/e^{1+\frac{k}{b}}) \cdot \OPTv$.
However, we must note that in practice, recomputing the vertex cover only tends to
improve the solution when the input exchange graph is sufficiently dense, and in
$2$-rendezvous where the actual
minimum vertex cover can be computed.

\begin{remark}
  \label{rem:anycomm} 
  According to \cite{otte2014any}, an algorithm is ``any-com'' if it finds ``a
  suboptimal solution quickly and refines it as communication permits''.
  The greedy algorithm described above has a similar trait:
  (i) for any budget $b$, it finds a near-optimal solution; (ii) let
  $(\Lcal_\text{grd}^b,\VV_\text{grd}^b)$ be the pair of near-optimal solutions produced for budget
  $b$. Then, $\VV_\text{grd}^b$ must be sent as a priority queue to robots to
  initiate the data exchange process by following the original ordering prescribed by the greedy algorithm (i.e.,
  first round, second round, etc). Now imagine the exchange process is
  interrupted after exchanging $b^\prime < b$ observations.
  Due to the incremental nature of the greedy algorithm, at this point robots
  have already exchanged $\VV_\text{grd}^{b^\prime}$, which is the solution that
  would have been produced by the greedy algorithm if the budget was
  $b^\prime$. But note that we know that this solution and its corresponding subset of loop closures
  $\Lcal_\text{grd}^{b^\prime}$ are near-optimal for budget $b^\prime$. 
\end{remark}

\noindent2) \underline{\it Non-uniform Observation Size}:
A modified version of the greedy algorithm is guaranteed
to provide a solution with an approximation factor of $1/2 \cdot (1-1/e)$
\cite{leskovec2007cost,krauseSurvey} (Table~\ref{tab:results}) for the knapsack
constraint. The algorithm is
very intuitive: first, we run the standard greedy algorithm
described above (stopping condition in this case will be the knapsack
constraint). Then, we rerun the greedy algorithm with a minor modification:
instead of picking the vertex with the highest marginal gain, we select the one
with the highest \emph{normalized} marginal gain,
i.e., $\big(\fv(\VV \cup \{v\}) - \fv(\VV)\big)/{w(v)}$. Finally, we return the
better solution. Note that the normalized marginal gain 
encodes the marginal gain achieved in our task-oriented objective 
per one bit of data transmission. For example, for $f_\text{NLC}$ \eqref{eq:NLC} this term
quantifies the expected number of realized loop closures gained by broadcasting
$v \in \Vall$ per unit of transmitted data.

\subsection{Computational Complexity Under Uniform Weights}
\label{subsec:complexity}
The standard greedy algorithm needs $O(b\cdot m)$ evaluations of
$\fv$ where $m$ denotes the number of vertices in the
exchange graph (e.g., number of images among which the search for inter-robot loop
closures takes place). Note that in practice $m \geq n$ depends on the number of
rendezvousing robots $n$. The number of function calls can be reduced using the
so-called lazy greedy method \cite{minoux1978accelerated,krauseSurvey}. Evaluating $f_\text{NLC}$
takes $O(1)$ time. By contrast, evaluating $f_\text{WST}$ and $f_\text{FIM}$
can be quite costly as they both require computing log-determinant of matrices of size $O(d)$ where
$d$ denotes the number robot poses in the global pose-graph. For dense
pose-graphs, a na\"{i}ve implementation runs in $O(b\cdot
m\cdot d^3)$ time. A more clever implementation
\cite{kasra16wafr,kasra18ijrr} can reduce
this to $O(b \cdot d^3 + b \cdot m \cdot d^2)$, or even $O(d^3 + b \cdot
m \cdot d^2)$. Leveraging the sparse structure
of real-world pose-graphs eliminates the cubic dependence of run time on $d$.
In addition, the greedy algorithm enjoys an ``embarrassingly parallel'' structure
that can be easily exploited to
significantly speed up the task of screening candidates and thus reduce the
impact of $m$ on the run time.

\subsection{Certifying Near-Optimality via Convex Relaxation}
\label{subsec:cvx}
Evaluating the performance of the proposed approximation algorithms for a
particular instance of our problem requires
the value of $\OPT$.
Computing $\OPT$
by brute force is impractical even in rather small scenarios. Given
an upper bound 
$\UPT \geq \OPT$, one can compare the
value attained by an approximation algorithm with $\mathrm{UPT}$
as a surrogate for $\mathrm{OPT}$. Therefore, $\UPT$ provides an \emph{a posteriori} certificate for 
near-optimality of $\VV_\text{grd}$.

We obtain such a certificate
by first
formulating our combinatorial optimization problem as an integer program (similar to the standard integer linear
programming formulation of the maximum
coverage problem), and then computing the optimal value of its natural
convex relaxation. 
Consider indicator variables $\ppp \triangleq [\pi_1,\dots,\pi_{n}]^\top$ and
$\boldsymbol{\ell}  \triangleq [\ell_1,\dots,\ell_m]^\top$ where each $\pi_i$
corresponds to a vertex (similar to \cite{Giamou18_ICRA}; see
Section~\ref{sec:problem_formulation}) and each $\ell_i$ corresponds to an edge
in the exchange graph. Let $\AAA$ be the (undirected) incidence matrix of the
exchange graph and $\mathbf{w}$ be the stacked vector of vertex weights. Define $\Fcal_{\text{int}} \triangleq \big\{(\ppp,\zz) \in \{0,1\}^n \times \{0,1\}^m
:\mathbf{w}^{\hspace{-0.05cm}\top}\ppp\leq
  b, \mathbf{A}^{\hspace{-0.1cm}\top}\ppp \geq \zz\big\}$. The objective
  functions introduced in Section~\ref{sec:problem_formulation} can all be
  expressed in terms of indicator variables $\zz$. For example, the D-criterion \eqref{eq:FIM}
  can be expressed as $h(\zz) \triangleq \log\det(\HH_\text{init} + \sum_{e \in
	\Lcal_\text{all}} \ell_e\cdot
  p(e) \cdot
  \HH_e)$; see also \cite{kasra16wafr,carlone2017attention}. It is easy to observe that maximizing $h(\zz)$ subject to $(\ppp,\zz) \in
  \Fcal_\text{int}$ is equivalent to \ref{eq:adapt} with the corresponding
  objective function. Now relaxing $\Fcal_\text{int}$ to 
  $\Fcal\triangleq \big\{(\ppp,\zz) \in [0,1]^n \times [0,1]^m
	:\mathbf{w}^{\hspace{-0.05cm}\top}\ppp\leq
  b, \mathbf{A}^{\hspace{-0.1cm}\top}\ppp \geq \zz\big\}$ results in convex
  optimization problems for each of the objective functions introduced in
  Section~\ref{sec:problem_formulation} (an LP for $f_\text{NLC}$ and
  max-det problems \cite{vandenberghe1998determinant} subject to additional
  affine constraints for $f_\text{WST}$ and $f_\text{FIM}$).
  This upper bound is used in Section~\ref{sec:experiments} to certify the
  near-optimality of the solutions produced by the greedy algorithms. These
  problem instance-specific certificates are often much stronger than our 
  theoretical worst-case guarantees.

\section{Experiments}
\label{sec:experiments}
In this section, we demonstrate the main features of our proposed approach with
both simulated and real-world datasets. In our experiments, all pose-graph SLAM
instances are solved using g2o \cite{kummerle2011g}. All convex relaxation
problems introduced in Section~\ref{subsec:cvx} are modeled using the YALMIP
toolbox \cite{Lofberg2004} and solved using SDPT3 \cite{Toh99sdpt3} in MATLAB.

\subsection{General Setup} 
A key property of our proposed approach is its
compatibility with any task-oriented NMS objective function. In our experiments, we consider the three functions introduced in Section~\ref{subsec:obj}: (i) D-optimality criterion (FIM); (ii) weighted tree-connectivity (WST), and (iii) expected number of loop closures (NLC).
Our main goal is not to compare these objective functions, but to highlight the
general applicability of our approach and evaluate it with standard
task-oriented metrics. Still, we point out that in practice, different
objectives lead to different computation and communication overhead in the
metadata exchange phase. As discussed in Section~\ref{subsec:complexity},
$f_\text{FIM}$ and $f_\text{WST}$ incur more computation overhead than
$f_\text{NLC}$, as evaluating them requires computing the log-determinant of
large matrices. In terms of communication, $f_{\text{FIM}}$ incurs the highest
overhead. $f_{\text{WST}}$ incurs less communication overhead than $f_{\text{FIM}}$, as it only requires the structure of the pose-graph \cite{kasra18ijrr}. Finally, $f_{\text{NLC}}$ incurs no communication overhead. 

In this work, we do not impose a communication budget on the metadata exchange
phase. This is done for two reasons. First, in the place recognition pipeline
introduced in Section~\ref{sec:problem_formulation}, metadata exchange only
constitutes a small fraction of the overall resource consumption. Second, the
overhead that arises in the metadata exchange phase is objective-dependent in
nature. Thus, counting the overhead as part of the resource consumption
introduces unnecessary bias towards certain objectives (e.g., $f_\text{NLC}$).
Nevertheless, we note that the proposed framework can be trivially extended to
account for additional resource consumption beyond the main data exchange phase.

We introduce two algorithms as baselines. 
One natural way to approach Problem~\ref{eq:adapt} is to greedily select edges $\Lcal$ from $\Eall$ until violating the vertex cover constraint 
$\vcover(\Lcal) \leq b$. In what follows, we call this baseline algorithm ``Edge
Greedy''. 
Note that Edge Greedy is a computationally expensive procedure, as it requires (approximately) solving the minimum vertex cover problem multiple times during the greedy loop. 
In our implementation, we further augment Edge Greedy by allowing it to select extra ``communication free'' edges, i.e., edges that are incident to the current vertex cover. 
Our second baseline is a simple random algorithm, which picks a random budget-feasible subset of observations and selects all potential loop closures that can be verified by sending these observations.

When comparing different algorithms, performance is evaluated in terms of the
normalized objective value, i.e. achieved objective value divided by the maximum
achievable value given infinite budget (i.e., selecting all candidates in $\Eall$). A crucial step is to demonstrate and verify the near-optimal performance of the proposed algorithm. Since finding the optimal
value of Problem~\ref{eq:adapt} is impractical,
we rely on the proposed convex relaxation approach (Section~\ref{subsec:cvx})
as an \emph{a posteriori} certificate of near-optimality. 
In addition, to quantify how much accuracy we lose by using only a subset of loop
closures, we compute the absolute trajectory error (ATE) between our
estimated trajectory and the full maximum likelihood estimate (obtained by
selecting all candidate loop closures). 

\subsection{KITTI Odometry Sequence 00}

\begin{figure}[t]
\centering

\begin{subfigure}[t]{0.2\textwidth}
\centering
\includegraphics[width=\textwidth]{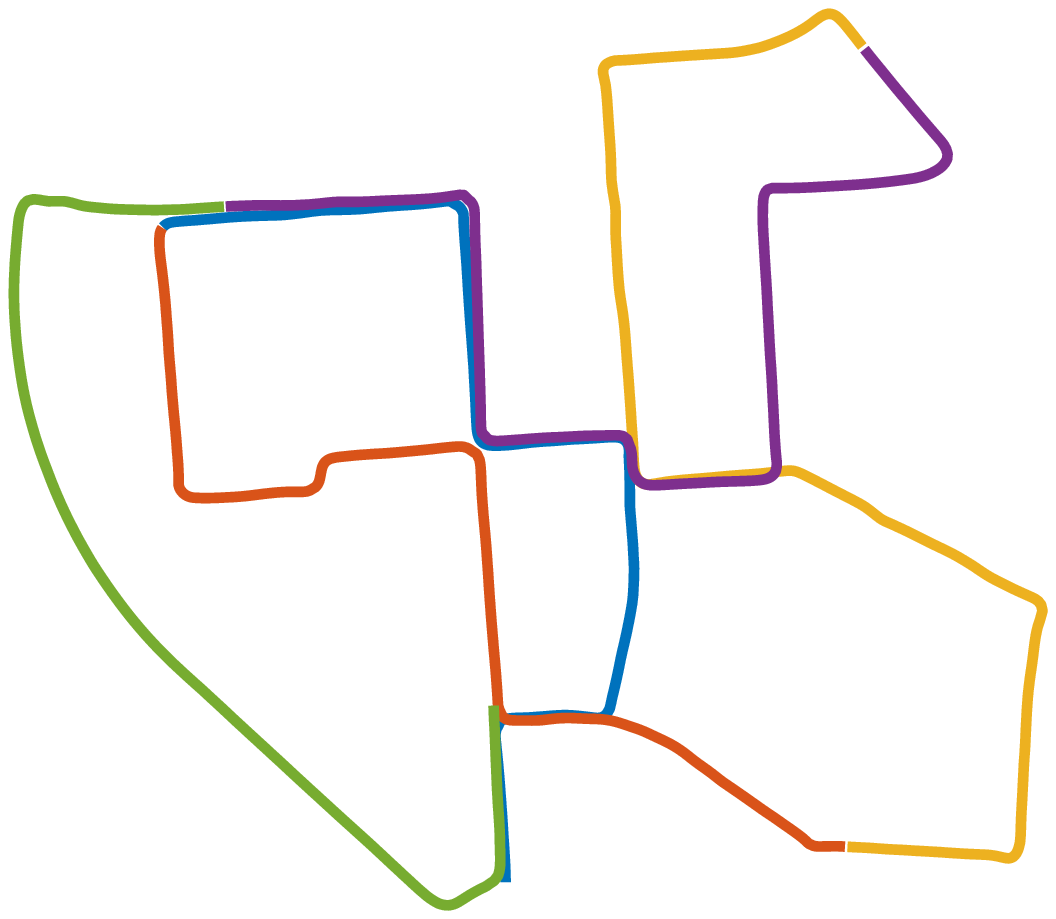}
\caption{\small KITTI 00}
\label{fig:KITTI_00_base_graph}
\end{subfigure}
~
\begin{subfigure}[t]{0.2\textwidth}
\centering
\includegraphics[width=\textwidth]{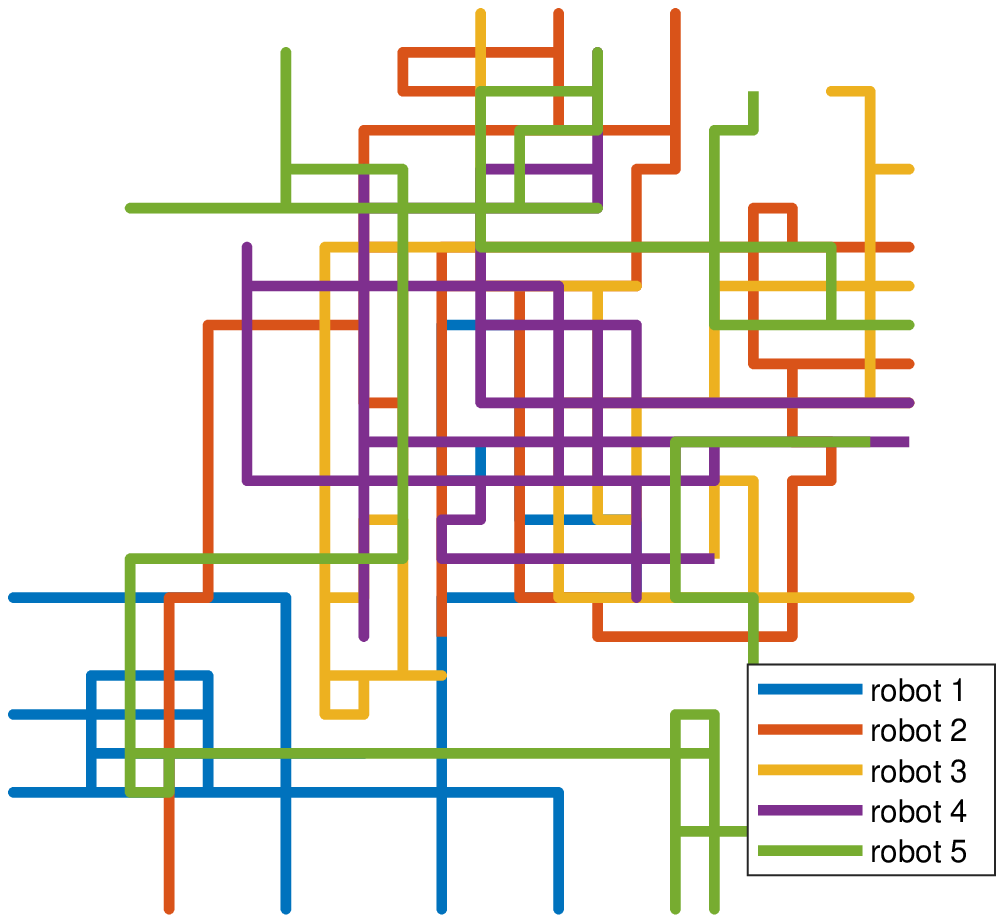}
\caption{\small Simulation}
\label{fig:Atlas_base_graph}
\end{subfigure}
\caption{\small Left: KITTI 00; Right: 2D simulation. Each base graph shows trajectories of five robots. Before inter-robot data exchange, trajectories are estimated purely using prior beliefs and odometry measurements, hence the drift displayed in the KITTI trajectories. The 
simulation trajectories shown are the exact ground truth.}
\end{figure}

In order to show that the proposed algorithms are useful in real-world
scenarios, we performed a number of offline experiments on data from the KITTI
odometry benchmark \cite{Geiger2012CVPR}. Similar to
\cite{CieslewskiChoudhary17}, we divide the odometry sequence $00$ into
sub-trajectories representing individual robots' paths (Figure
\ref{fig:KITTI_00_base_graph}). This particular sequence contains many realistic
path intersections and re-traversals, providing a large number of inter-robot
loop closures that make the measurement selection problem interesting. For the
KITTI experiments, we use a modified version of ORB-SLAM2 \cite{murORB2} to
generate stereo visual odometry and potential loop closures. As in
\cite{Giamou18_ICRA}, we define the probability associated with each potential
loop closure by normalizing the corresponding visual similarity score outputted
by DBoW2 \cite{GalvezTRO12}.  Since the weighted tree-connectivity (WST)
objective only supports 2D data, we project the KITTI data to the $x$-$y$ plane
and work with the projected 2D trajectories. The same projection is done for the
other two objective functions to ensure a fair comparison.

\begin{figure*}[t]
  \begin{subfigure}[t]{0.23\textwidth}
	\centering
	\includegraphics[width=\textwidth]{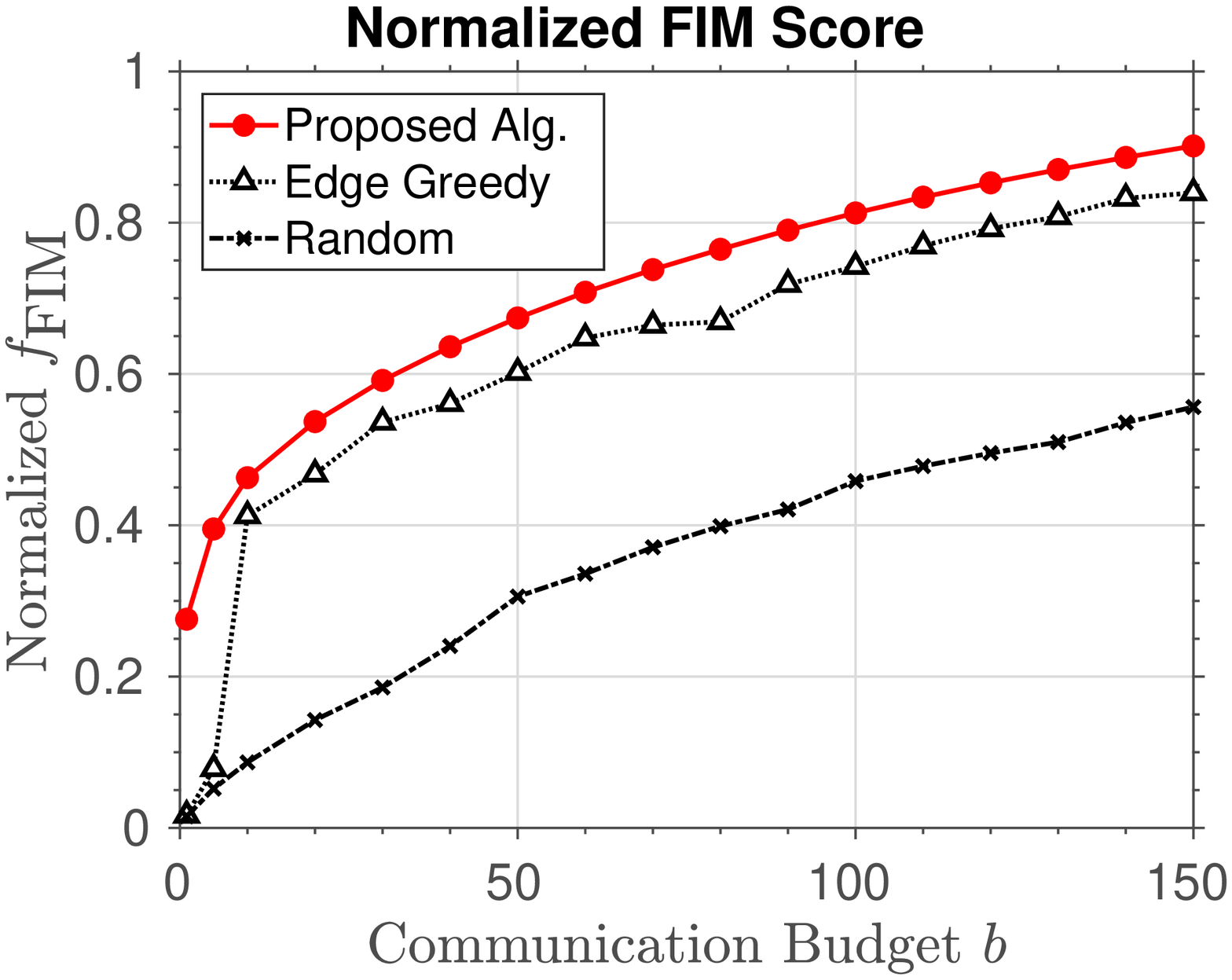}
	\caption{\small FIM (Eq. \ref{eq:FIM})}
	\label{fig:results_full_KITTI_00_FIM}
  \end{subfigure}
  ~
  \begin{subfigure}[t]{0.23\textwidth}
	\centering
	\includegraphics[width=\textwidth]{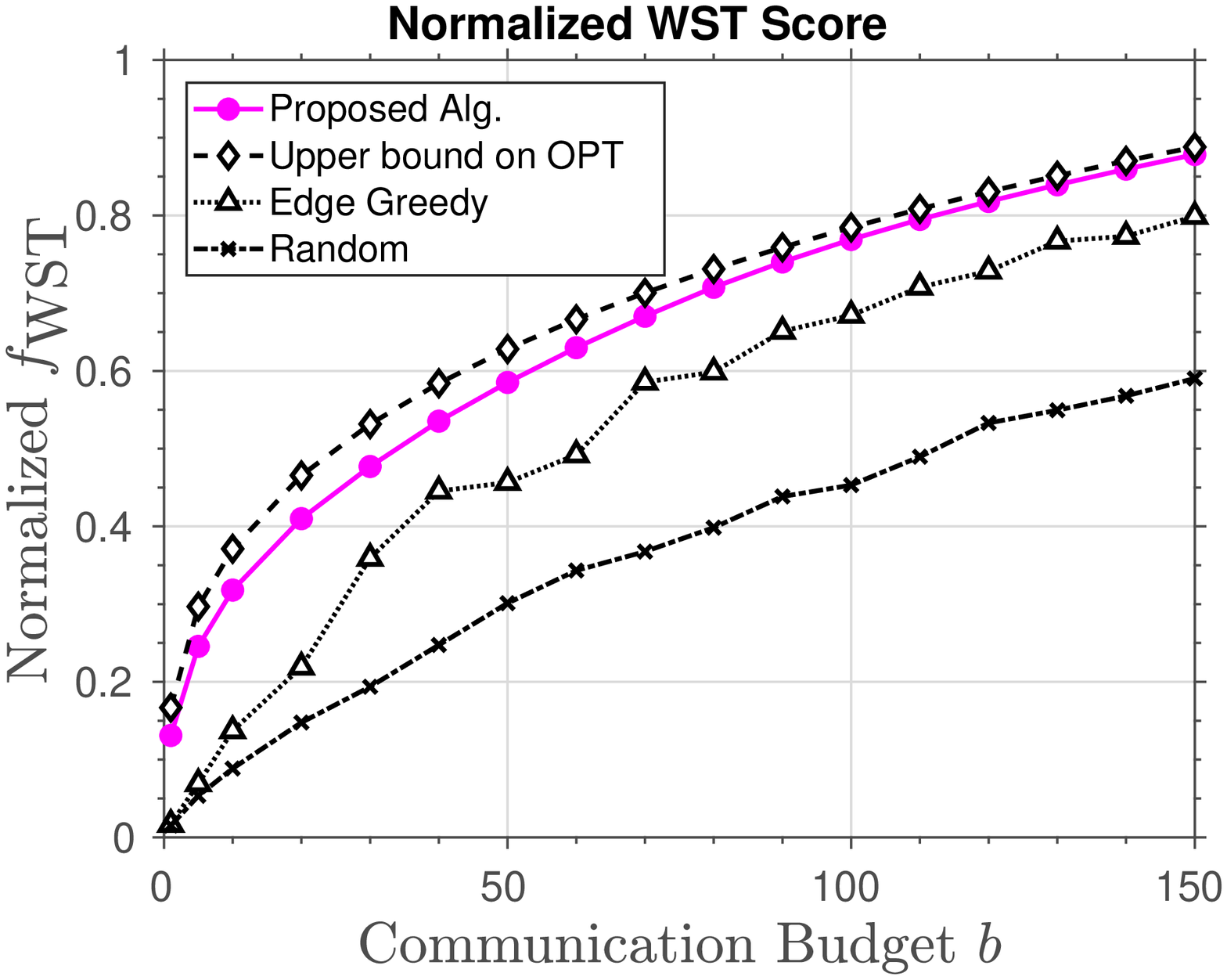}
	\caption{\small WST (Eq. \ref{eq:WST})}
	\label{fig:results_full_KITTI_00_WST}
  \end{subfigure}
  ~
  \begin{subfigure}[t]{0.23\textwidth}
	\centering
	\includegraphics[width=\textwidth]{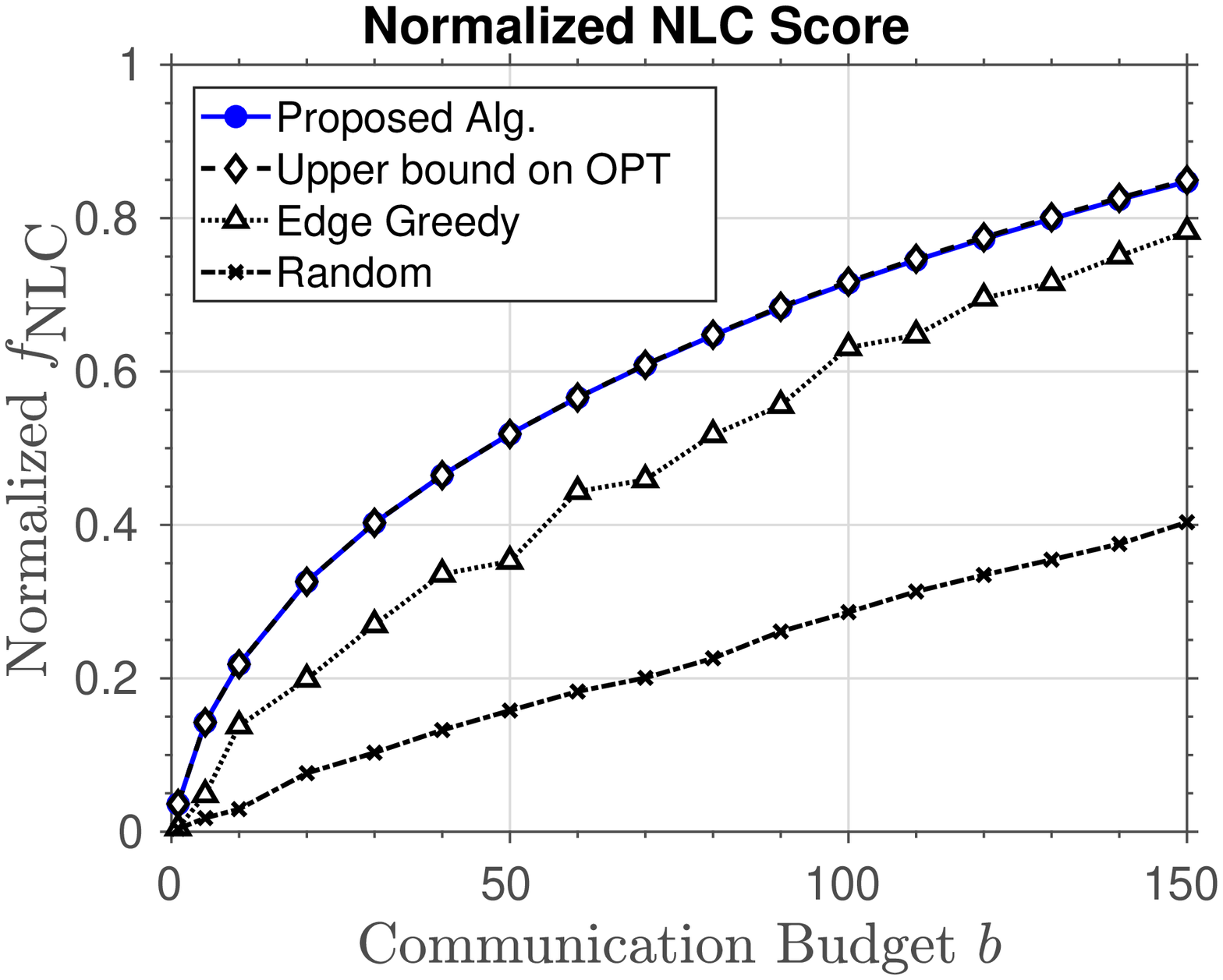}
	\caption{\small NLC (Eq. \ref{eq:NLC})}
	\label{fig:results_full_KITTI_00_LC}
  \end{subfigure}
  ~
\begin{subfigure}[t]{0.23\textwidth}
\centering
\includegraphics[width=\textwidth]{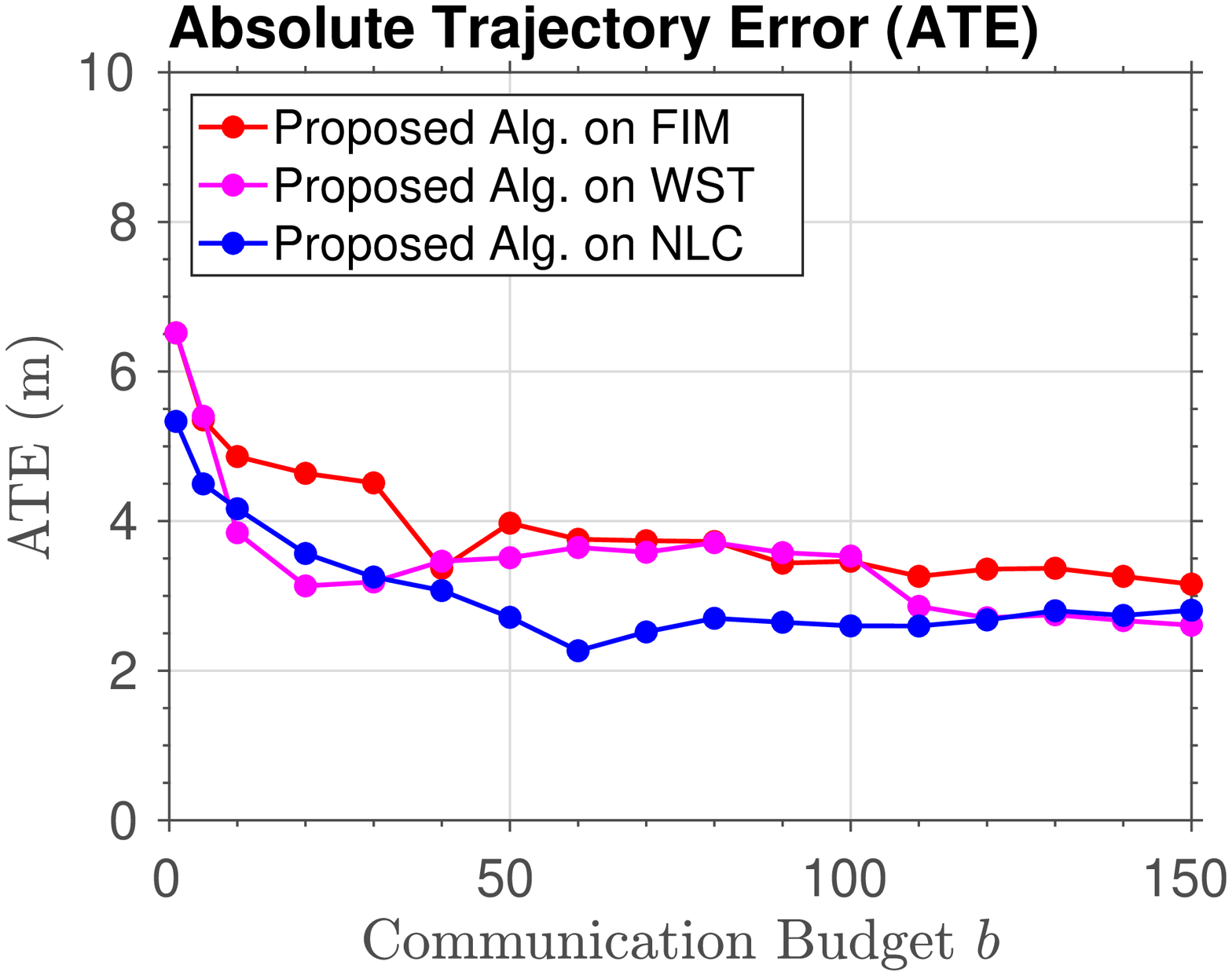}
\caption{\small ATE }
\label{fig:results_full_KITTI_00_realized}
\end{subfigure}
\caption{\small (a) to (c) shows performance of the proposed algorithm in KITTI 00 under
different objective functions and varying communication budget. Objective value is normalized by the maximum achievable value of each objective given infinite budget (selecting all potential loop closures). (d): cross-objective performance evaluated on ATE, compared against the maximum likelihood estimate given infinite budget. In the KITTI experiments, budget $b$ is defined as the total number of observations the team can exchange. In this dataset, the minimum budget to cover all potential loop closures is 250, which translate to roughly 20MB of total data exchange bandwidth.}
\label{fig:results_full_KITTI_00}
\vspace{-0.3cm}
\end{figure*}

In practice, the number of keypoints is roughly the same (around 2000) for all observations obtained by ORB-SLAM2. In our experiments, we ignore this insignificant variation in observation sizes and assign each vertex a unit weight. In this way, our communication budget reduces to a cardinality constraint on the total number of observations the team can exchange. 
Assuming each keypoint (consisting of a descriptor and coordinates) 
uses approximately 40 bytes of data, a communication budget of 50, for example, translates to 4MB of total data exchange bandwidth.

Figures \ref{fig:results_full_KITTI_00_FIM}-\ref{fig:results_full_KITTI_00_LC}
illustrate the performance of the proposed algorithm under
different objective functions and varying communication budget.
Under all three objectives, the proposed algorithm outperforms both
baselines by a significant margin. 
Intuitively, the Edge Greedy baseline blindly maximizes the objective function
without controlling resource consumption, while the random baseline only aims at
achieving a budget-feasible solution. In contrast, by having foresight over resource consumption while maximizing the objective, the proposed method spends the budget more wisely, hence always achieving a higher score. 
Furthermore, Figures~\ref{fig:results_full_KITTI_00_WST} (WST) and
\ref{fig:results_full_KITTI_00_LC} (NLC) show upper bounds on the value of the
optimal solution obtained using convex relaxation (Section~\ref{subsec:cvx}). 
We do not include the upper
bound for the FIM objective because solving the convex relaxation in this case
is too time consuming.\footnote{The KITTI dataset we use contains more than 2000
poses in total. Consequently, the Fisher information matrix with three degrees of
freedom has dimension larger than $6000 \times 6000$.} For both WST and NLC objectives, we
observe that the performance of the proposed algorithm is close to the convex
relaxation upper bound. This confirms our intuition that in practice, the proposed algorithm performs much better than the theoretical lower bound of $(1-1/e) \cdot \OPTe$.
 It is particularly interesting that for the NLC objective, the value achieved by the proposed approach matches exactly with the upper bound, indicating that in
this case the solution is actually optimal. 
This happens because
the specific problem instance is sparse, i.e., candidate edges form several
distinct connected components.
In this case, the entire problem breaks down into
multiple easier sub-problems, for which it is more likely for the greedy
algorithm to obtain an optimal solution. 
Due to the same sparse structure, we do not observe improvement caused by recomputing the minimum vertex cover, 
as discussed in Section~\ref{subsec:apxalg}.

Figure \ref{fig:results_full_KITTI_00_realized} illustrates the cross-objective
performance of the proposed method evaluated on 
the absolute trajectory error (ATE).  
We note that the overall decrease in error is small as communication budget increases. This suggests that our initial estimate is relatively accurate. Nevertheless, as the communication budget increases, the estimates of
all three objectives eventually approach the full maximum likelihood estimate (obtained by selecting all potential loop closures). 
Interestingly,
the ATE metric is not monotonically decreasing. Furthermore, the performance of the NLC
objective is more stable compared to the FIM and WST objectives. Both observations are due to the fact that 
the ATE is highly dependent 
on the random realization of the selected loop closures. 
Intuitively, although the FIM and WST objectives tend to select more informative loop closure candidates,
these candidates may turn out to be false positives and hence do not contribute to the final localization accuracy.
From this perspective, the NLC objective is advantageous as it seeks 
to maximize the expected number of true loop closures. 
However, the other two objectives can be easily augmented
to better cope with the stochastic nature of candidate loop closures.
This can be done
by, e.g., prefiltering the candidates and removing those 
with low occurrence probabilities.

\subsection{Simulation Experiments}
\begin{figure}[t]
\centering
\includegraphics[width=0.32\textwidth]{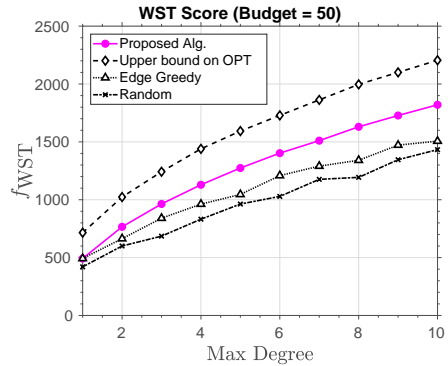}
\caption{\small Performance of the proposed algorithm maximizing the WST objective in simulation under fixed communication budget (50 in this case) and varying maximum degree of the exchange graph.}
\label{fig:wst_maxdeg_budget_50}
\vspace{-0.2cm}
\end{figure}

Synthetic data was produced using the 2D simulator functionality of g2o \cite{kummerle2011g}. The simulator produced noisy odometry and loop closure constraints for robots moving in a random square grid pattern (Figure
\ref{fig:Atlas_base_graph}). The trajectory was divided into multiple robots in the 
same manner as the KITTI data, but loop closures probabilities $p$ were generated from a uniform random distribution. 
We use a simulated graph to study the behavior of the proposed algorithm under varying densities of the input exchange graph.
In our simulation, we control this density by enforcing different maximum degrees on the vertices and randomly pruning excess candidate edges. 
Figure \ref{fig:wst_maxdeg_budget_50} displays the effect of increasing density (in terms of the maximum vertex degree) on the WST objective. 
The algorithms were applied with a fixed communication budget of 50 over the same exchange graph. As expected, the proposed algorithm outperforms the baselines by a significant margin. 

\section{Conclusion \& Future Work}
\label{sec:conclusion}
Inter-robot loop closures constitute the essence of collaborative localization
and mapping; without them ``collaboration'' is impossible. Detecting
inter-robot loop closures is a resource-intensive process, during which robots
must exchange substantial amounts of sensory data. Recent works have made great
strides in making CSLAM front-ends more data efficient
\cite{CieslewskiChoudhary17,Giamou18_ICRA}. While such approaches play a crucial role
in saving scarce mission-critical resources, in many scenarios the resources
available onboard (mainly, battery and bandwidth) may be insufficient for
verifying every potential loop closure. Consequently, robots must be able to solve
the inter-robot loop closure detection problem under budgeted communication.

This paper addressed this critical challenge by adopting a resource-adaptive
approach. In particular, we presented ``any-com'' constant-factor approximation
algorithms for the problem of selecting a budget-feasible subset of potential loop closures
while maximizing a task-oriented objective.  Performance guarantees presented in
this work hold for any monotone submodular objective.  Extensive experimental
results using the KITTI benchmark dataset and realistic synthetic data validated
our theoretical results.

We plan to extend our results by considering new forms of communications constraints.
In particular, in heterogeneous multirobot systems, one may wish to partition the team into
several disjoint blocks and impose different communication budgets on each
block. Our preliminary results indicate that our approach can be extended to
cover this case. While beyond the scope of
this paper, analyzing more complex communication protocols and network
topologies is another important future direction. 

\section*{Acknowledgement}
This work was supported in part by the NASA Convergent Aeronautics Solutions
project Design Environment for Novel Vertical Lift Vehicles (DELIVER), by the
Northrop Grumman Corporation, and by ONR under BRC award
N000141712072.

\clearpage
\bibliographystyle{plainnat}
\bibliography{scanexchange}

\begin{thebibliography}{40}
\providecommand{\natexlab}[1]{#1}
\providecommand{\url}[1]{\texttt{#1}}
\expandafter\ifx\csname urlstyle\endcsname\relax
  \providecommand{\doi}[1]{doi: #1}\else
  \providecommand{\doi}{doi: \begingroup \urlstyle{rm}\Url}\fi

\bibitem[Ageev and Sviridenko(1999)]{ageevGraphMaxCover}
Alexander~A. Ageev and Maxim~I. Sviridenko.
\newblock Approximation algorithms for maximum coverage and max cut with given
  sizes of parts.
\newblock In G{\'e}rard Cornu{\'e}jols, Rainer~E. Burkard, and Gerhard~J.
  Woeginger, editors, \emph{Integer Programming and Combinatorial
  Optimization}, pages 17--30, Berlin, Heidelberg, 1999. pringer Berlin
  Heidelberg.
\newblock ISBN 978-3-540-48777-7.

\bibitem[Ausiello et~al.(2012)Ausiello, Crescenzi, Gambosi, Kann,
  Marchetti-Spaccamela, and Protasi]{ausiello2012complexity}
Giorgio Ausiello, Pierluigi Crescenzi, Giorgio Gambosi, Viggo Kann, Alberto
  Marchetti-Spaccamela, and Marco Protasi.
\newblock \emph{Complexity and approximation: Combinatorial optimization
  problems and their approximability properties}.
\newblock Springer Science \& Business Media, 2012.

\bibitem[Carlone and Karaman(2017)]{carlone2017attention}
Luca Carlone and Sertac Karaman.
\newblock Attention and anticipation in fast visual-inertial navigation.
\newblock In \emph{Robotics and Automation (ICRA), 2017 IEEE International
  Conference on}, pages 3886--3893. IEEE, 2017.

\bibitem[Caskurlu et~al.(2014)Caskurlu, Mkrtchyan, Parekh, and
  Subramani]{caskurluBipartiteMaxCover}
Bugra Caskurlu, Vahan Mkrtchyan, Ojas Parekh, and K.~Subramani.
\newblock On partial vertex cover and budgeted maximum coverage problems in
  bipartite graphs.
\newblock In Josep Diaz, Ivan Lanese, and Davide Sangiorgi, editors,
  \emph{Theoretical Computer Science}, pages 13--26, Berlin, Heidelberg, 2014.
  Springer Berlin Heidelberg.

\bibitem[Choudhary et~al.(2017)Choudhary, Carlone, Nieto, Rogers, Christensen,
  and Dellaert]{choudhary2017}
Siddharth Choudhary, Luca Carlone, Carlos Nieto, John Rogers, Henrik~I
  Christensen, and Frank Dellaert.
\newblock Distributed mapping with privacy and communication constraints:
  Lightweight algorithms and object-based models.
\newblock \emph{The International Journal of Robotics Research}, 36\penalty0
  (12):\penalty0 1286--1311, 2017.
\newblock \doi{10.1177/0278364917732640}.

\bibitem[Cieslewski and Scaramuzza(2017{\natexlab{a}})]{CieslewskiS17}
Titus Cieslewski and Davide Scaramuzza.
\newblock Efficient decentralized visual place recognition from full-image
  descriptors.
\newblock In \emph{1st International Symposium on Multi-Robot and Multi-Agent
  Systems}, 2017{\natexlab{a}}.

\bibitem[Cieslewski and
  Scaramuzza(2017{\natexlab{b}})]{cieslewski2017efficient}
Titus Cieslewski and Davide Scaramuzza.
\newblock Efficient decentralized visual place recognition using a distributed
  inverted index.
\newblock \emph{IEEE Robotics and Automation Letters}, 2\penalty0 (2):\penalty0
  640--647, 2017{\natexlab{b}}.

\bibitem[Cieslewski et~al.(2017)Cieslewski, Choudhary, and
  Scaramuzza]{CieslewskiChoudhary17}
Titus Cieslewski, Siddharth Choudhary, and Davide Scaramuzza.
\newblock Data-efficient decentralized visual {SLAM}.
\newblock \emph{CoRR}, abs/1710.05772, 2017.
\newblock URL \url{http://arxiv.org/abs/1710.05772}.

\bibitem[Crescenzi(1997)]{crescenzi1997short}
Pierluigi Crescenzi.
\newblock A short guide to approximation preserving reductions.
\newblock In \emph{Computational Complexity, 1997. Proceedings., Twelfth Annual
  IEEE Conference on (Formerly: Structure in Complexity Theory Conference)},
  pages 262--273. IEEE, 1997.

\bibitem[G\'alvez-L\'opez and Tard\'os(2012)]{GalvezTRO12}
Dorian G\'alvez-L\'opez and J.~D. Tard\'os.
\newblock Bags of binary words for fast place recognition in image sequences.
\newblock \emph{IEEE Transactions on Robotics}, 28\penalty0 (5):\penalty0
  1188--1197, October 2012.
\newblock ISSN 1552-3098.
\newblock \doi{10.1109/TRO.2012.2197158}.

\bibitem[Geiger et~al.(2012)Geiger, Lenz, and Urtasun]{Geiger2012CVPR}
Andreas Geiger, Philip Lenz, and Raquel Urtasun.
\newblock Are we ready for autonomous driving? the kitti vision benchmark
  suite.
\newblock In \emph{Conference on Computer Vision and Pattern Recognition
  (CVPR)}, 2012.

\bibitem[Giamou et~al.(2018)Giamou, Khosoussi, and How]{Giamou18_ICRA}
Matthew Giamou, Kasra Khosoussi, and Jonathan~P How.
\newblock Talk resource-efficiently to me: Optimal communication planning for
  distributed loop closure detection.
\newblock In \emph{IEEE International Conference on Robotics and Automation
  (ICRA)}, 2018.

\bibitem[Hochbaum(1996)]{hochbaum1996approximation}
Dorit~S Hochbaum.
\newblock \emph{Approximation algorithms for NP-hard problems}.
\newblock PWS Publishing Co., 1996.

\bibitem[Huang et~al.(2013)Huang, Kaess, and Leonard]{Huang2013}
G.~Huang, M.~Kaess, and J.~J. Leonard.
\newblock Consistent sparsification for graph optimization.
\newblock In \emph{2013 European Conference on Mobile Robots}, pages 150--157,
  Sept 2013.
\newblock \doi{10.1109/ECMR.2013.6698835}.

\bibitem[Joshi and Boyd(2009)]{Joshi2009}
Siddharth Joshi and Stephen Boyd.
\newblock Sensor selection via convex optimization.
\newblock \emph{Signal Processing, IEEE Transactions on}, 57\penalty0
  (2):\penalty0 451--462, 2009.

\bibitem[Karp(1972)]{karp1972reducibility}
Richard~M Karp.
\newblock Reducibility among combinatorial problems.
\newblock In \emph{Complexity of computer computations}, pages 85--103.
  Springer, 1972.

\bibitem[Khosoussi et~al.(2016{\natexlab{a}})Khosoussi, Huang, and
  Dissanayake]{kasra16icra}
Kasra Khosoussi, Shoudong Huang, and Gamini Dissanayake.
\newblock Tree-connectivity: Evaluating the graphical structure of {SLAM}.
\newblock In \emph{Robotics and Automation (ICRA), 2016 IEEE International
  Conference on}, pages 1316--1322. IEEE, 2016{\natexlab{a}}.

\bibitem[Khosoussi et~al.(2016{\natexlab{b}})Khosoussi, Sukhatme, Huang, and
  Dissanayake]{kasra16wafr}
Kasra Khosoussi, Gaurav~S. Sukhatme, Shoudong Huang, and Gamini Dissanayake.
\newblock Designing sparse reliable pose-graph {SLAM}: A graph-theoretic
  approach.
\newblock \emph{International Workshop on the Algorithmic Foundations of
  Robotics}, 2016{\natexlab{b}}.

\bibitem[Khosoussi et~al.(2018)Khosoussi, Giamou, Sukhatme, Huang, Dissanayake,
  and How]{kasra18ijrr}
Kasra Khosoussi, Matthew Giamou, Gaurav~S Sukhatme, Shoudong Huang, Gamini
  Dissanayake, and Jonathan~P How.
\newblock Reliable graph topologies for {SLAM}.
\newblock \emph{International Journal of Robotics Research}, 2018.
\newblock Accepted.

\bibitem[Krause and Golovin(2014)]{krauseSurvey}
Andreas Krause and Daniel Golovin.
\newblock Submodular function maximization.
\newblock In Lucas Bordeaux, Youssef Hamadi, and Pushmeet Kohli, editors,
  \emph{Tractability: Practical Approaches to Hard Problems}, pages 71--104.
  Cambridge University Press, 2014.
\newblock ISBN 9781139177801.

\bibitem[Kretzschmar and Stachniss(2012)]{kretzschmar2012}
Henrik Kretzschmar and Cyrill Stachniss.
\newblock Information-theoretic compression of pose graphs for laser-based
  slam.
\newblock \emph{The International Journal of Robotics Research}, 31\penalty0
  (11):\penalty0 1219--1230, 2012.
\newblock \doi{10.1177/0278364912455072}.
\newblock URL \url{https://doi.org/10.1177/0278364912455072}.

\bibitem[K{\"u}mmerle et~al.(2011)K{\"u}mmerle, Grisetti, Strasdat, Konolige,
  and Burgard]{kummerle2011g}
Rainer K{\"u}mmerle, Giorgio Grisetti, Hauke Strasdat, Kurt Konolige, and
  Wolfram Burgard.
\newblock g 2 o: A general framework for graph optimization.
\newblock In \emph{Robotics and Automation (ICRA), 2011 IEEE International
  Conference on}, pages 3607--3613. IEEE, 2011.

\bibitem[Leonardos et~al.(2017)Leonardos, Zhou, and
  Daniilidis]{leonardos2017distributed}
Spyridon Leonardos, Xiaowei Zhou, and Kostas Daniilidis.
\newblock Distributed consistent data association via permutation
  synchronization.
\newblock In \emph{Robotics and Automation (ICRA), 2017 IEEE International
  Conference on}, pages 2645--2652. IEEE, 2017.

\bibitem[Leskovec et~al.(2007)Leskovec, Krause, Guestrin, Faloutsos,
  VanBriesen, and Glance]{leskovec2007cost}
Jure Leskovec, Andreas Krause, Carlos Guestrin, Christos Faloutsos, Jeanne
  VanBriesen, and Natalie Glance.
\newblock Cost-effective outbreak detection in networks.
\newblock In \emph{Proceedings of the 13th ACM SIGKDD international conference
  on Knowledge discovery and data mining}, pages 420--429. ACM, 2007.

\bibitem[L{\"{o}}fberg(2004)]{Lofberg2004}
J.~L{\"{o}}fberg.
\newblock Yalmip : A toolbox for modeling and optimization in matlab.
\newblock In \emph{In Proceedings of the CACSD Conference}, Taipei, Taiwan,
  2004.

\bibitem[Minoux(1978)]{minoux1978accelerated}
Michel Minoux.
\newblock Accelerated greedy algorithms for maximizing submodular set
  functions.
\newblock In \emph{Optimization techniques}, pages 234--243. Springer, 1978.

\bibitem[Mur-Artal and Tard\'os(2017)]{murORB2}
Ra\'ul Mur-Artal and Juan~D. Tard\'os.
\newblock {ORB-SLAM2}: an open-source {SLAM} system for monocular, stereo and
  {RGB-D} cameras.
\newblock \emph{IEEE Transactions on Robotics}, 33\penalty0 (5):\penalty0
  1255--1262, 2017.
\newblock \doi{10.1109/TRO.2017.2705103}.

\bibitem[Nemhauser et~al.(1978)Nemhauser, Wolsey, and
  Fisher]{nemhauser1978analysis}
George~L Nemhauser, Laurence~A Wolsey, and Marshall~L Fisher.
\newblock An analysis of approximations for maximizing submodular set
  functions—i.
\newblock \emph{Mathematical Programming}, 14\penalty0 (1):\penalty0 265--294,
  1978.

\bibitem[Nerurkar et~al.(2009)Nerurkar, Roumeliotis, and
  Martinelli]{nerurkar2009distributed}
Esha~D Nerurkar, Stergios~I Roumeliotis, and Agostino Martinelli.
\newblock Distributed maximum a posteriori estimation for multi-robot
  cooperative localization.
\newblock In \emph{Robotics and Automation, 2009. ICRA'09. IEEE International
  Conference on}, pages 1402--1409. IEEE, 2009.

\bibitem[Otte and Correll(2014)]{otte2014any}
Michael Otte and Nikolaus Correll.
\newblock Any-com multi-robot path-planning with dynamic teams: Multi-robot
  coordination under communication constraints.
\newblock In \emph{Experimental Robotics}, pages 743--757. Springer, 2014.

\bibitem[Paull et~al.(2015)Paull, Huang, Seto, and
  Leonard]{paull2015communication}
Liam Paull, Guoquan Huang, Mae Seto, and John~J Leonard.
\newblock Communication-constrained multi-auv cooperative {SLAM}.
\newblock In \emph{Robotics and Automation (ICRA), 2015 IEEE International
  Conference on}, pages 509--516. IEEE, 2015.

\bibitem[Paull et~al.(2016)Paull, Huang, and Leonard]{paull2016unified}
Liam Paull, Guoquan Huang, and John~J Leonard.
\newblock A unified resource-constrained framework for graph {SLAM}.
\newblock In \emph{Robotics and Automation (ICRA), 2016 IEEE International
  Conference on}, pages 1346--1353. IEEE, 2016.

\bibitem[Pukelsheim(1993)]{Pukelsheim1993}
Friedrich Pukelsheim.
\newblock \emph{Optimal design of experiments}, volume~50.
\newblock SIAM, 1993.

\bibitem[Saeedi et~al.(2016)Saeedi, Trentini, Seto, and Li]{saeedi2016multiple}
Sajad Saeedi, Michael Trentini, Mae Seto, and Howard Li.
\newblock Multiple-robot simultaneous localization and mapping: A review.
\newblock \emph{Journal of Field Robotics}, 33\penalty0 (1):\penalty0 3--46,
  2016.

\bibitem[Shamaiah et~al.(2010)Shamaiah, Banerjee, and
  Vikalo]{shamaiah2010greedy}
Manohar Shamaiah, Siddhartha Banerjee, and Haris Vikalo.
\newblock Greedy sensor selection: Leveraging submodularity.
\newblock In \emph{49th IEEE Conference on Decision and Control (CDC)}, pages
  2572--2577. IEEE, 2010.

\bibitem[Sviridenko(2004)]{sviridenko2004note}
Maxim Sviridenko.
\newblock A note on maximizing a submodular set function subject to a knapsack
  constraint.
\newblock \emph{Operations Research Letters}, 32\penalty0 (1):\penalty0 41--43,
  2004.

\bibitem[Toh et~al.(1999)Toh, Todd, and Tütüncü]{Toh99sdpt3}
K.~C. Toh, M.J. Todd, and R.~H. Tütüncü.
\newblock {SDPT3} -- a matlab software package for semidefinite programming.
\newblock \emph{Optimization Methods and Software}, 11:\penalty0 545--581,
  1999.

\bibitem[Vandenberghe et~al.(1998)Vandenberghe, Boyd, and
  Wu]{vandenberghe1998determinant}
Lieven Vandenberghe, Stephen Boyd, and Shao-Po Wu.
\newblock Determinant maximization with linear matrix inequality constraints.
\newblock \emph{SIAM journal on matrix analysis and applications}, 19\penalty0
  (2):\penalty0 499--533, 1998.

\bibitem[Vazirani(2013)]{vazirani2013approximation}
Vijay~V Vazirani.
\newblock \emph{Approximation algorithms}.
\newblock Springer Science \& Business Media, 2013.

\bibitem[Vial et~al.(2011)Vial, Durrant-Whyte, and Bailey]{Vial2011}
J.~Vial, H.~Durrant-Whyte, and T.~Bailey.
\newblock Conservative sparsification for efficient and consistent approximate
  estimation.
\newblock In \emph{2011 IEEE/RSJ International Conference on Intelligent Robots
  and Systems}, pages 886--893, Sept 2011.
\newblock \doi{10.1109/IROS.2011.6095128}.

\end{thebibliography}

\clearpage
\pagebreak
\appendix

\begin{algorithm}[t]
  \caption{Greedy Algorithm (Uniform Observation Sizes)}\label{alg:vgreedy}
  \begin{algorithmic}[1]
	\renewcommand{\algorithmicrequire}{\textbf{Input:}}
	\renewcommand{\algorithmicensure}{\textbf{Output:}}
	\Require
	  \Statex - Exchange graph $\Gcal = (\Vall, \Eall)$
	  \Statex - Data transmission budget $b$ 
	  \Statex - $\fe: 2^{\Eall} \to \Rset_{\geq 0}$ and 
	  $\fv : \VV \mapsto
	    \fe\big(\edg(\VV)\big)$
	\Ensure
	\Statex - A near-optimal budget-feasible pair 
	$\Vgrd \subseteq \Vall, \Egrd \subseteq \Eall$. 
	Robots then exchange observations in $\Vgrd$ and collectively verify the potential loop closures in $\Egrd$.
	\vspace{0.08cm}
	\Function{VertexGreedy}{$\Gcal$,$\fv$,$b$}
	\State {$\Vgrd \leftarrow \varnothing, \Egrd \leftarrow \varnothing, \Ccal \leftarrow \varnothing,  \cost \leftarrow 0$}
	\While {\textsc{True}}
	\State \textcolor{green!50!black}{// keep track of results from previous round}
	\State $\Vprev \leftarrow \Vgrd, \mathcal{C}_\text{prev} \leftarrow \mathcal{C}$
	\While{$\cost < b$} \Comment \textcolor{green!50!black}{greedy loop}
		\State {$v^\star \leftarrow \argmax_{v \in
		  \Vall\setminus\Vgrd} \fv(\Vgrd \cup \{v\})$}
			\State {$\Vgrd \leftarrow \Vgrd \cup \{v^\star\}$}
			\State {$\cost \leftarrow \cost + 1$}
		\EndWhile
		\State $\Egrd \leftarrow \edg(\Vgrd)$ 
		\State \textcolor{green!50!black}{// vertex cover provides certificate of feasibility}
		\State {$\mathcal{C} \leftarrow \textsc{ApxVertexCover}(\Egrd)$}\footnotemark \label{line:vcover1} 
		\State $\cost \leftarrow \min(\cost, |\mathcal{C}| )$
		\If {$\cost = b$}
		\State \textbf{break}
		\EndIf
	\EndWhile
	\State $\Vnew \leftarrow \Vgrd/\Vprev$ \Comment \textcolor{green!50!black}{selected in the last round}
	\State $\Vgrd \leftarrow \mathcal{C}_\text{prev} \cup \Vnew$ \Comment \textcolor{green!50!black}{budget feasible}
	\State $\Egrd \leftarrow \edg(\Vgrd)$ 
	\State \Return $\Vgrd, \Egrd$
	\EndFunction
  \end{algorithmic}
\end{algorithm}
\footnotetext{
  In $2$-rendezvous, \textsc{ApxVertexCover} returns a minimum vertex cover.
In general $n$-rendezvous ($n \geq 3$), the returned solution is a $2$-approximation
(i.e., the size of the resulting vertex cover is
at most twice the size of the minimum vertex cover) \cite{vazirani2013approximation}. 
\label{footnote}}

\begin{algorithm}[t]
  \caption{\textsc{ApxVertexCover} \cite{vazirani2013approximation}}
  \begin{algorithmic}[1]
	\renewcommand{\algorithmicrequire}{\textbf{Input:}}
	\renewcommand{\algorithmicensure}{\textbf{Output:}}
	\Require
	  \Statex - Edge set $\Ecal$
	\Ensure
	\Statex - A vertex cover $\Ccal \subseteq \VV(\Ecal)$ for the graph induced
	by $\Ecal$ such that $|\Ccal|$ is at most twice the size of a minimum vertex cover.
	If the induced graph is bipartite, the returned vertex cover is a minimum vertex cover.
	\vspace{0.2cm}
	\Function{ApxVertexCover}{$\Ecal$}
	\State Solve the LP relaxation of the minimum vertex cover
	\[
	  \ppp^\star \in \underset{\ppp}{\argmin} \,\, \mathbf{1}^{\hspace{-0.05cm}\top}\ppp \,\,
	  \text{s.t.} \,\, \mathbf{A}^{\hspace{-0.05cm}\top}\ppp \geq \mathbf{1},
	\ppp \geq \mathbf{0}.\]
	$\mathbf{A}$ is the incidence matrix of the graph induced by
	$\Ecal$.
	\State \textcolor{green!50!black}{// round the LP solution}
	\State $\Ccal \leftarrow \{v \in \VV(\Ecal):\ppp_v^\star \geq 1/2 \}$
	\State \Return $\Ccal$
	\EndFunction
  \end{algorithmic}
\end{algorithm}

\subsection{Proofs}
\begin{proof}[\bfseries Proof of Theorem~\ref{thm:fe2fv}]

\begin{itemize}
\item Normalized:
\begin{align}
\fv(\varnothing) & = \fe(\edg(\varnothing)) &&\ptext{by def.
\eqref{eq:fvdef}}\\&=
\fe(\varnothing) && \ptext{by def. of $\edg$} \\&= 0 && \ptext{$\fe$ is normalized}.
\label{}
\end{align}

\item Monotone:
we need to show that $\fv(\Scal \cup\{v\})
 \geq
\fv(\Scal)$ for any $\Scal \subseteq \Vall$ and $v \in \Vall$. This holds since,
\begin{align}
\fv(\Scal \cup \{v\}) & = \fe\big(\edg(\Scal \cup \{v\})\big) &&
\ptext{by def. \eqref{eq:fvdef}}\\
& \geq \fe\big(\edg(\Scal)\big) && \ptext{$\fe$ is monotone.}\\
& = \fv(\Scal) && \ptext{by def. \eqref{eq:fvdef}}.
\label{}
\end{align}

\item Submodular: We need to show the following holds
 for any $\Scal \subseteq \Vall$ and $\Qcal \subseteq \Vall$:
 \begin{align}
   \fv(\Scal) + \fv(\Qcal) \geq \fv(\Scal\cup\Qcal) + \fv(\Scal\cap\Qcal)
   \label{}
 \end{align}
 or, equivalently,
 \begin{align}
   \fe\big(\edg(\Scal)\big) & + \fe\big(\edg(\Qcal)\big) \geq \nonumber \\
   \fe\big(\edg(\Scal \cup \Qcal)\big) & + \fe\big(\edg(\Scal \cap \Qcal)\big).
   \label{}
 \end{align}
 Now since $\fe$ is submodular we have,
  \begin{align}
    \fe\big(\edg(\Scal)\big) & + \fe\big(\edg(\Qcal)\big) \geq \nonumber \\
    \fe\big(\edg(\Scal) \cup \edg(\Qcal)\big) & + \fe\big(\edg(\Scal) \cap \edg(\Qcal)\big).
    \label{}
  \end{align}
  Therefore, it suffices to show that,
  \begin{align}
    \fe\big(\edg(\Scal) \cup \edg(\Qcal)\big) & + \fe\big(\edg(\Scal) \cap
    \edg(\Qcal)\big) \geq \nonumber \\
    \fe\big(\edg(\Scal \cup \Qcal)\big) & + \fe\big(\edg(\Scal \cap \Qcal)\big).
    \label{}
  \end{align}
  Lemma~\ref{lem:cup} below shows that,
   \begin{equation}
     \fe\big(\edg(\Scal)  \cup \edg(\Qcal)\big) =  \fe\big(\edg(\Scal \cup
     \Qcal)\big).\label{eq:cup}
   \end{equation}
  It only remains to show the following inequality holds:
   \begin{align}
     \fe\big(\edg(\Scal)  \cap \edg(\Qcal)\big) &\geq  \fe\big(\edg(\Scal \cap
     \Qcal)\big).\label{eq:cap}
   \end{align}
   The above inequality follows from Lemma~\ref{lem:cap} and the fact that
   $\fe$ is monotone. This concludes the proof.
\end{itemize}
\end{proof}
\vspace{0.5cm}

\begin{lemma} $\edg(\Scal)  \cup \edg(\Qcal) = \edg(\Scal \cup \Qcal)$.
  \label{lem:cup}
\end{lemma}
\vspace{0.2cm}
\begin{proof}[\bfseries Proof of Lemma~\ref{lem:cup}]
     We first show that,
	 \begin{equation}
	   \edg(\Scal) \cup \edg(\Qcal) \subseteq \edg(\Scal \cup \Qcal).
	   \label{eq:dir1}
	 \end{equation}
	 Consider any edge $e \in \edg(\Scal) \cup \edg(\Qcal)$.
	 There must be a vertex $v$ in $\Scal$ and/or in $\Qcal$
	 that is incident to $e$. Thus $v \in \Scal \cup \Qcal$, and consequently, $e \in \edg(\Scal \cup \Qcal)$.
	 This proves \eqref{eq:dir1}.

	 Now we show that
	 \begin{equation}
	   \edg(\Scal) \cup \edg(\Qcal) \supseteq \edg(\Scal \cup \Qcal).
	   \label{eq:dir2}
	 \end{equation}
	 This time let $\tilde{e} \in \edg(\Scal \cup \Qcal)$. Hence, there must be
	 a vertex $\tilde{v}$ either in $\Scal$ and/or $\Qcal$ such that it is
	 incident to
	 $\tilde{e}$. Therefore, $\tilde{e} \in \edg(\Scal) \cup \edg(\Qcal)$. This
	 proves \eqref{eq:dir2}, and concludes the proof of this lemma.
\end{proof}
\vspace{0.5cm}
\begin{lemma}
   $\edg(\Scal)  \cap \edg(\Qcal) \supseteq  \edg(\Scal \cap
   \Qcal)$.
  \label{lem:cap}
\end{lemma}
\vspace{0.2cm}
\begin{proof}[\bfseries Proof of Lemma~\ref{lem:cap}]
 Let $e \in \edg(\Scal \cap \Qcal)$. There must be a vertex $v \in \Scal \cap
 \Qcal$ incident to $e$. Now note that $e \in \edg(\Scal)$ and $e \in \edg(\Qcal)$; thus
 $e \in \edg(\Scal) \cap \edg(\Qcal)$. This concludes the proof.
\end{proof}
\vspace{0.5cm}

\begin{proof}[\bfseries Proof of Lemma~\ref{th:opts}]
\begin{enumerate}
\item From the fact that $\tilde{\VV}$ is \ref{prob:pv}-feasible we have $\sum_{v
	\in \tilde{\VV}} w(v) \leq b$.
Now note that $\edg(\tilde{\VV})$ by definition is covered by $\tilde{\VV}$.
We therefore have,
\begin{align}
	\vcover\big(\edg(\tilde{\VV})\big) & = \sum_{\mathclap{v \in
 \Vcover(\edg(\tilde{\VV}))}} w(v) && \ptext{by def. of
	$\vcover$}\\ & \leq  \sum_{v \in
 \tilde{\VV}} w(v) && \ptext{$\edg(\tilde{\VV})$ is covered by
 $\tilde{\VV}$} \\ & \leq b.
 &&\ptext{$\tilde{\VV}$ is \ref{prob:pv}-feasible}
\end{align}

\item We show $\OPTv = \OPTe$ by first showing $\OPTv \leq \OPTe$ and then $\OPTv
  \geq \OPTe$.\\[0.2cm]
  I) \textbf{[$\boldsymbol{\leq}$]:} Let $\VV^\star \subseteq \VV$ be an
  optimal solution for \ref{prob:pv}. From the first statement in this lemma
  we already know that $\edg(\VV^\star)$
  is \ref{eq:adapt}-feasible since $\VV^\star$ is \ref{prob:pv}-feasible.
  Now we have,
  \begin{align}
	\OPTv & \triangleq \fv(\VV^\star) && \ptext{by def.} \\
	& = \fe\big(\edg(\VV^\star)\big) && \ptext{by def. of $\fv$}\\
	& \leq \OPTe. && \ptext{$\edg(\VV^\star)$ is \ref{eq:adapt}-feasible}
	\label{}
  \end{align}\\[0.2cm]
  II) \textbf{[$\boldsymbol{\geq}$]:}
  Let $\Lcal^\star \subseteq \EE$
  be an optimal solution for \ref{eq:adapt}. This implies that $\Lcal^\star$ is \ref{eq:adapt}-feasible; i.e., $\vcover(\Lcal^\star) \leq
  b$.
  It is easy to observe that $\Vcover(\Lcal^\star)$
  is \ref{prob:pv}-feasible:
  \begin{align}
	\sum_{\mathclap{v \in \Vcover(\Lcal^\star)}} w(v) = \vcover(\Lcal^\star)
	\leq b.
	\label{}
  \end{align}
  Furthermore, note that $\Lcal^\star \subseteq
  \edg\big(\Vcover(\Lcal^\star)\big)$. Then we have:
  \begin{align}
	\fe(\Lcal^\star) & \leq \fe\big(\edg(\Vcover(\Lcal^\star))\big) &&
	\ptext{$\fe$ is monotone}\\
	& \leq \fv\big(\Vcover(\Lcal^\star)\big). && \ptext{by def. of $\fv$}
	\label{eq:importantIneq}
  \end{align}
  Using this result we have,
  \begin{align}
	\OPTe & \triangleq \fe(\Lcal^\star) && \ptext{by def.}\\
	& \leq \fv\big(\Vcover(\Lcal^\star)\big) &&
	\ptext{follows from \eqref{eq:importantIneq}}\\
	& \leq \OPTv. &&
	\ptext{$\Vcover(\Lcal^\star)$ is \ref{prob:pv}-feasible}
	\label{}
  \end{align}
  This completes the proof.
\end{enumerate}
\end{proof}
\vspace{0.5cm}
\begin{proof}[\bfseries Proof of Theorem~\ref{th:alphaalg}]
  \begin{align}
	\fe\big(\edg(\VV)\big) & = \fv(\VV) && \ptext{by def. of $\fv$}\\
	& \geq \alpha \cdot \OPTv && \ptext{assumption}\\
	& = \alpha \cdot \OPTe. && \ptext{Theorem~\ref{th:opts}}.
	\label{}
  \end{align}
\end{proof}

\subsection{Extension to Theorem~\ref{thm:fe2fv}}
We presented Theorem~\ref{thm:fe2fv} in the context of an exchange graph $\Gcal = (\Vall, \Eall)$.
Nevertheless, this result can be easily extended to a more general setting. 
Let $\Ucal$ be a universe of elements and $\Scal$ be a collection of subsets of $\Ucal$.
We note that our exchange graph is a special case of this general formulation,
as $\Eall$ can be treated as the universe of elements and 
$\Vall$ as the collection of subsets.\footnote{
The same insight also arises in the well known reduction from the vertex cover problem to the set cover problem.}
Now consider an NMS function over the element set $\Ucal$:
\begin{equation}
f_\Ucal: 2^{\Ucal} \to \Rset_{\geq 0}.
\end{equation}
Similar to (\ref{eq:fvdef}), we can define a corresponding function over the collection of subsets $\Scal$:
\begin{equation}
f_\Scal : 2^{\Scal} \to \Rset_{\geq 0 }: \Wcal \mapsto
  f_\Ucal\big(\mathsf{elements}(\Wcal)\big).
\end{equation}
where $\mathsf{elements} : 2^{\Scal} \to 2^{\Ucal}$ gives the set of all elements 
contained by at least one selected subset.
Then, we have the following generalized theorem.
\begin{theorem}
  \normalfont
  For any NMS $f_\Ucal$, the corresponding $f_\Scal$ is NMS.
  \label{thm:fe2fv_general}
\end{theorem}
Theorem~\ref{thm:fe2fv_general} can be proved in a similar way as Theorem~\ref{thm:fe2fv}. The detailed proof is omitted.

\end{document}